%% file: main.tex
\documentclass[10pt,twocolumn,letterpaper]{article}

\newif\ifdebug
\debugtrue % comment out before submitting!

\newif\ifsqueeze
\squeezetrue

\input{macros}

\iccvfinalcopy % *** Uncomment this line for the final submission

\def\iccvPaperID{2} % *** Enter the ICCV Paper ID here

\ificcvfinal\fi
\begin{document}
	
	%%%%%%%%% TITLE
	\title{Efficient Real-Time Camera Based Estimation of Heart Rate and Its Variability {\small $\heartsuit$}}%\hspace{0.2942cm}
	%\title{Towards Efficient real-time Estimation of Heart Rate \& its Variability {\small $\heartsuit$} }
	%$\overrightarrow{\text{Towards}}$ Efficient and real-time Heart Rate (Variability) Estimation $\heartsuit$
	%\\\todo{Choose better title}
		%Towards Efficiency in rPPG: Real-time Heart Rate and Variability from Webcams 
	%\jvg{not a fan of the title. I would remove "Towards Efficiency in rPPG:". Also 'variability' is not clear that this 'heart rate variability' is meant. How about: "Towards Real-time Heart rate variability estimation from webcams", sure normal heart rate is not included, but that is implied. } } 
	
	\author{
		{\hspace{3.2cm}}
		\and
		Amogh Gudi\textsuperscript{ {\large$\star$} VV TUD}\\
		%Vicarious Perception Technologies (VicarVision), Delft University of Technology \\
		{\tt\small amogh@vicarvision.nl}
		% \and
		% {.\hspace{1cm}.}
		\and
		Marian Bittner\textsuperscript{ {\large$\star$} VV TUD}\\ % \$
		%VicarVision, TU Delft \\
		{\tt\small marian@vicarvision.nl}
		\and
		{\hspace{3.2cm}}
		\and
		%{}
		\and
		Roelof Lochmans\textsuperscript{ TU/e}\\
		%VicarVision, TU Delft \\
		{\tt\small r.h.lochmans@student.tue.nl}
		\and
		%{.\hspace{1cm}.}
		%\and
		Jan van Gemert\textsuperscript{ TUD}\\
		%Delft University of Technology \\
		{\tt\small j.c.vangemert@tudelft.nl}
		\and
		%{}
		\and
		{\hspace{-0.248cm}\normalsize \textsuperscript{VV }Vicarious Perception Technologies\hspace{-0.248cm}}\\
		{\small Amsterdam, The Netherlands}\\
		\and
		{\hspace{-0.248cm}\normalsize \textsuperscript{TUD }Delft University of Technology\hspace{-0.248cm}}\\
		{\small Delft, The Netherlands}\\
		\and
		{\hspace{-0.248cm}\normalsize \textsuperscript{TU/e }Eindhoven University of Technology\hspace{-0.248cm}}\\
		{\small Eindhoven, The Netherlands}\\
		%\and
		%{\rule{10cm}{0.0005cm}}\\
		%{\footnotesize \textsuperscript{\normalsize $\star$} Equal contribution.}\\
	}
	
	\maketitle
	\ificcvfinal\blfootnote{\kern-2.2em \textsuperscript{\normalsize$\star$} Equal contribution.}\fi
	%\ificcvfinal\blfootnote{\kern-2.2em \textsuperscript{{\footnotesize\$}} Partially funded by the CAS Project G-1234-1234-1234.}\fi
	\input{abstract}
	\input{introduction}
	\input{relatedwork}
	\input{method}
	\input{experiments}
	\input{conclusion}
	{\small
		\bibliographystyle{ieee}
		\bibliography{../references}
	}
	%\clearpage
	%\bibliographystyle{ieee}
	%\bibliography{../references}
	%\nocite{*}
\end{document}

%% file: macros.tex
\usepackage{iccv}

\usepackage{booktabs}
\usepackage{blindtext}
\usepackage{color}
\usepackage{soul}
\soulregister\cite7
\soulregister\ref7
\soulregister\pageref7
\usepackage[linesnumbered,lined,ruled,vlined]{algorithm2e}
\SetAlFnt{\scriptsize}

\SetCommentSty{mycommfont}
\usepackage[font=small, skip=0pt]{caption} %uncomment to make captions smaller!
\usepackage{subfig}
\usepackage{epstopdf} %Enables .eps image graphics
\usepackage{floatrow}
\usepackage[percent]{overpic}
\usepackage[table]{xcolor}
\usepackage{tabularx}
\usepackage{setspace}
\usepackage{float}
\usepackage{times}
\usepackage{textcomp}
\usepackage{epsfig}
\usepackage{graphicx}
\graphicspath {{Figures/}}
\usepackage{amsmath}
\usepackage{amssymb}
\usepackage{wrapfig}
\usepackage{bibentry}
\usepackage[inline]{enumitem}
\usepackage[normalem]{ulem}
\usepackage[symbol]{footmisc}
\renewcommand{\thefootnote}{$\star$}
\newcommand\blfootnote[1]{%
  \begingroup
  \renewcommand\thefootnote{}\footnote{#1}%
  \addtocounter{footnote}{-1}%
  \endgroup
}
\nonstopmode

%% file: abstract.tex
\begin{abstract}
%\noindent
\label{sec:abstract}
Remote photo-plethysmography (rPPG) uses a remotely placed camera to estimating a person's heart rate (HR).
Similar to how heart rate can provide useful information about a person's vital signs, insights about the underlying physio/psychological conditions can be obtained from heart rate variability (HRV). 
HRV is a measure of the fine fluctuations in the intervals between heart beats.
However, this measure requires temporally locating heart beats with a high degree of precision.
We introduce a refined and efficient real-time rPPG pipeline with novel filtering and motion suppression that not only estimates heart rate more accurately, but also extracts the pulse waveform to time heart beats and measure heart rate variability.
This method requires no rPPG specific training and is able to operate in real-time.
We validate our method on a self-recorded dataset under an idealized lab setting, and show state-of-the-art results on two public dataset with realistic conditions (VicarPPG and PURE).
\end{abstract}

% (a measure of `heart skipping a beat').
%\hl{(One of)} 
%\jvg{Remove: 'The task of '} \jvg{Try to end the sentence with HR because your next sentence starts with HR: 'Remote photo-plethysmography (rPPG) involves estimating a person's heart rate (HR) via a remotely placed camera.' = 'Remote photo-plethysmography (rPPG) uses a remotely placed camera to estimating a person's heart rate (HR).'}  % by observing changes in skin reflectance.
%While heart rate is an important vital sign measure, a

%While HR is an important vital sign,  \jvg{dont make it a competition: 'another measure of higher utility is the heart rate variability (HRV):' = 'Important extra information is in heart rate variability (HRV):'} fine fluctuations of the intervals between heart beats (a measure of `heart skipping a beat').
%Thus, unlike HR estimation which is an aggregation operation where errors can cancel out, in HRV analysis even small artefacts add up and strongly distort the measurement. This makes this task extra challenging to perform via rPPG.
%\hl{This is one of the first work to do so}.

% by relying on appearance modelling based facial skin pixel selection and a two-stage FFT filtering and noise reduction. \todo{break this into smaller sentences}

%\jvg{no competition: remove 'even surpassing a deep learning-set benchmark'.}.

%% file: introduction.tex
\section{Introduction}
\label{sec:intro}
%\todo{better opening statement?}
Human vital signs like heart rate, blood oxygen saturation and related physiological measures can be measured using a technique called photo-plethysmography (PPG). This technique involves optically monitoring light absorption in tissues that are associated with blood volume changes.
%Photo-plethysmography (PPG) is the technique of optically monitoring light absorption in tissue that are associated with blood volume changes. 
%\jvg{try to avoid acronyms, they take mental effort from reader so if BVP never comes back in the intro, don't mention the acronym}
%These changes are often used to monitor vital signs like heart rate, blood oxygen saturation and related physiological measures.
Typically, this is done via a contact sensor attached to the skin surface \cite{allen2007photoplethysmography}.
%\jvg{If you say 'typically' I expect at least one but maybe even 2 or 3 citations.}.
%essentially performs the same operation
%\jvg{do not use reference words (see writing guidelines) better to repeat what you refer to with 'same operation' as it may be unclear what is referred to, while absorbing mental effort from the reader}
Remote Photo-plethysmography (rPPG) detects the blood volume pulse remotely by tracking changes in the skin reflectance as observed by a camera \cite{hassan2017heart,verkruysse2008remote}. 
%\mb{Should we cite the first rPPG paper here or better an overview?}\ag{both}
In this paper we present a novel framework for extracting heart rate (HR) and heart rate variability (HRV) from the face.
 %\jvg{space}
 %\jvg{add a space: 'rate(HR)' = 'rate (HR)' }
%\jvg{'Face': Why Capital 'F' in Face? Royal Persons Only? Or Face Of God?} in real-time.
%\hl{In this paper, we demonstrate a method that performs rPPG on faces to extract heart rate and heart rate variability in real-time.}

The process of rPPG essentially involves two steps: detecting and tracking the skin colour changes of the subject, and analysing this signal to compute measures like heart rate, heart rate variability and respiration rate. 
Recent advances in computer video, signal processing, and machine learning have improved the performances of rPPG techniques significantly \cite{hassan2017heart}.
Current state-of-the-art methods are able to leverage image processing by deep neural networks to robustly select skin pixels within an image and perform HR estimation \cite{chen2018deepphys,vspetlikvisual}. 
%\jvg{So? What is your point? Conclusion missing.}
\begin{figure*}
	\centering
	\includegraphics[width=0.9\textwidth, trim={0 0.55cm 0 0}, clip]{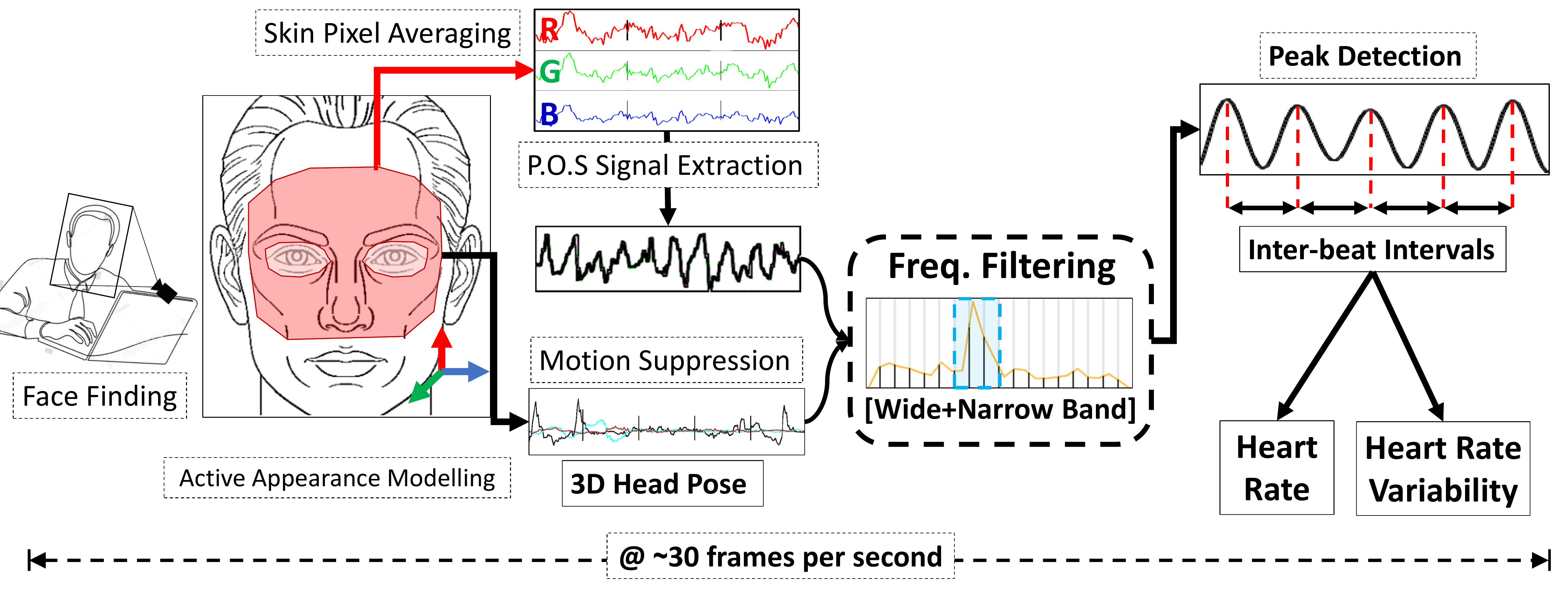}
	\caption{An overview of the proposed heart rate and heart rate variability estimation pipeline (left to right). 
	The face in captured webcam images are detected and modelled to track the skin pixels in region of interest. A single 1-D signal is extracted from the spatially averaged values of these pixels over time. 
	In parallel, 3-D head movements are tracked and used to suppress motion noise. 
	An FFT based wide and narrow band filtering process produces a clean pulse waveform from which peaks are detected. 
	The inter-beat intervals obtained from these peaks are then used to compute heart rate and heart rate variability.
	The full analysis can be performed in real time on a CPU.
	}
	%\jvg{make sure caption fully explain the fig: a reader first goes into 'comic book' mode. Add a conclusion to each caption, if you can.} 
	\label{fig:overview}
	%\vspace{-8pt}
\end{figure*}
However, 
%\jvg{Never start a paragraph with 'however' see my writing guidelines} 
%\jvg{It seems this belongs with prev. paragraph?} 
this reliance upon heavy machine learning (ML) processes has two primary drawbacks:
\begin{enumerate*}[label=(\roman*)]
	\item it necessitates rPPG specific training of the ML model, thereby requiring collection of large training sets; 
	%\jvg{remove 'explicit', or.. as opposed to implicit?}
	\item complex models can require significant computation time on CPUs and thus can potentially add a bottleneck in the pipeline and limit real-time utility. 
	%\jvg{Is that true? CNNs are quite efficient at inference time}
\end{enumerate*}
Since rPPG analysis is originally a signal processing task, the use of an end-to-end trainable system with no domain knowledge leaves room for improvement in efficiency (e.g., we \textit{know} that pulse signal is embedded in average skin colour changes \cite{verkruysse2008remote, Wu12Eulerian, zhang2017video}, but the ML system has to \textit{learn} this).
We introduce a simplified and efficient rPPG pipeline that performs the full rPPG analysis in real-time. This method achieves state-of-the-art results without needing any rPPG related training.
%We introduce a refined and efficient facial appearance modelling based rPPG pipeline that performs the full rPPG analysis in real-time without needing any explicit rPPG specific training.
This is achieved via extracting regions of interest robustly by 3D face modelling, and explicitly reducing the influence of head movement to filter the signal.
%\hl{This is achieved via a faster active appearance modelling (AAM) based skip pixel selection, and a novel wide/narrow band FFT filtering and motion noise suppression.}
%\mb{This is achieved by modelling the face to track regions of interest robustly over time, extracting color information,reducing influence of head motion on the resulting signal and filtering it. }
%\jvg{Text in yellow cannot be understood at this point in the text. Can you 'dumb it down' (ie: make it slightly high-level) so it can be understood? }

While heart rate is a useful output from a PPG/rPPG analysis, finer analysis of the obtained blood volume pulse (BVP) signal can yield further useful measures. 
One such measure is heart rate variability: an estimate of the variations in the time-intervals between individual heart beats. This measure has high utility in providing insights into physiological and psychological state of a person (stress levels, anxiety, etc.). While traditionally this measure is obtained based on observation over hours, short and ultra-short duration ($\scriptstyle\le$ 5 mins) HRV are also being studied \cite{shaffer2017overview}.
Our experiments focus on obtaining ultra-short HRV measure as a proof-of-concept/technology demonstrator for longer duration applications.
%the short and
%\jvg{I like it.}

The computation of heart rate variability requires temporally locating heart beats with a high degree of accuracy.
%\jvg{avoid reference words such as 'this measure': spell it out, see my writing guidelines} 
Unlike HR estimation, where errors in opposite directions average out, HRV analysis is sensitive to even small artefacts and all errors add up to strongly distort the final measurement.
% which is a dominant frequency selection / aggregation operation
%\hl{(equally strong errors in opposite directions result is zero average error)}
%In comparison, the task of average HR estimation is simpler: simply selecting the dominant frequency bin in the spectrum can be sufficient.
Thus, estimating HRV is a challenging task for rPPG and this has received relatively little focus in literature.
% most current methods do not attempt this.
%Thus, estimating HRV is a challenging task for rPPG and most current methods do not attempt this. \mb{Tom says: and there have recently been no methods that attemptt this?}
Our method extracts a clean BVP signal from the input via a two step wide and narrow band frequency filter to accurately time heart beats and estimate heart rate variability.
%\jvg{good, but can you lift the veil slightly on how? E.g.: mention that you explicitly remove head pose and other things? }
%This is one of the first methods to demonstrate this via webcam based rPPG. \todo{too strong?}

\ifsqueeze\vspace{-0.3cm}\fi
\paragraph{Contributions}
We make the following contributions in this work:
\begin{enumerate*}[label=(\bf \roman*)]
	\item We present a refined and efficient rPPG pipeline that can estimate heart-rate with state-of-the-art accuracy from RGB webcams. 
	This method has the advantage that it does not require any specific rPPG training and it can perform its analysis with real-time speeds.
	\item Our method is able to time individual heart beats in the estimated pulse signal to compute heart rate variability. 
	This body of work has received little attention, and we set the first benchmarks on two publicly available datasets.
	%\mb{Tom says: the first?} 
	\item We provide an in-depth HR and HRV estimation analysis of our method on a self-recorded dataset as well as publicly available datasets with realistic conditions (VicarPPG~\cite{tasli2014remote}, PURE~\cite{stricker2014non}, MAHNOB-HCI~\cite{soleymani2012amultimodal}).
	%\jvg{add citations to each data set} 
	We show state-of-the-art results on VicarPPG and PURE. 
	This also surpasses a previous benchmark set by a deep learning based method on PURE.
	%\mb{\underline{should} we put emphasis on that?}
	%\jvg{i'll allow the muscle talk, for this one time.. :) }
\end{enumerate*}

%% file: relatedwork.tex
\section{Related Work}
\label{sec:relwork}

% \paragraph{Origins of rPPG}
% %\todo{PPG method [1 paper only]; and first/basic rPPG method [poh \etal, EVM, VV] -> shortly say how it works. (BSS based methods)}
% Photopleytismography (PPG) is a non-invasive optical technique that can be used to measure heart rate, oxygen saturation and related physical measures.
% PPG works by measuring the absorption of light of certain wavelengths in tissue. 
% Varying amounts of blood in blood vessels traversing the skin are associated with changes in the absorbed light.\cite{allen2007photoplethysmography}
% PPG devices normally use one or multiple light sources that emit light at certain wavelengths to measure these changes, however,
% it was shown that a digital camera can be a suitable sensor to collect this information in ambient light conditions as well. \cite{verkruysse2008remote}
% \todo{ Add to start of paper}

% - Photopleytismography is a non-invasive optical technique hr and other physiological parameters
% - Recent years have seen an increase interest in PPG developments (Cite Sun paper)
% - Measuring changes in the blood volume pulsing through capillaries. (Cite Hassan paper)
% - Gaining popularity in the last years is PPG using a camera sensor. 
% - Camera sensors are cheap and general purpose. 
% - First paper using recordings at ambient light: Valkrysse. 

% Their works form the basis/foundation of the current state of the art in video based rPPG.
%recent works have further taken this forward and accuracies have improved.

\paragraph{Signal processing based rPPG methods}
%\todo{Mention rPPG good methods only based on signal processing. (remember that not all are real-time, could also mention this)}
% -Multiple algorithms since then.
% -Shown improvements
% -Two subtypes Blind source separation (ICA, PCA) and filter based
% -Main focus for us on non-blind source techniques (have problems to select a signal described in \todo{Cite Hassan})
% -Most prominent techniques
%     - Li background estimation \todo{Cite Li}
%     - Tuchyakov Self adaptive matrix \todo{Cite Tuchyakov}
%     - de Haan CHROMA,2SR, POS \todo{Cite de Haan/ Wang}
% Multiple techniques have been developed since then to improve the quality of the extracted BVP or BVP signal.
% The developed techniques can be divided into techniques that depend on a form of blind source separation e.g. (ICA, PCA) and techniques that don't.
% We will focus for the rest of this paragraph on the latter kind, since the developed algorithm does not make use of blind source separation.
%Multiple papers have been written on methods to extract heart rate or the BVP \jvg{Yes, here you can use 'blood volume (BVP)' } from videos.
Since the early work of Verkruysse \etal \cite{verkruysse2008remote},  who showed that heart rate could be measured from recordings from a consumer grade camera in ambient light, a large body of research has been conducted on the topic.
% a multitude of teams have worked to advance the field.} 
%\todo{replace multitude of teams maybe? Should we cite them all here?}
%Verkruysse \etal \cite{verkruysse2008remote} were the first to show that heart rate could be measured from recordings of a consumer grade camera in ambient light. 
Extensive reviews of these work can be found in \cite{hassan2017heart,rouast2018remote,sun2015photoplethysmography}.
Most published rPPG methods work either by applying skin detection on a certain area in each frame or by selecting one or multiple regions of interest and track their averages over time to generate colour signals. 
A general division can be made into methods that use blind source separation (ICA, PCA) \cite{mcduff2014improvements,mcduff2014remote,poh2010advancements} vs those that use a `fixed' extraction scheme for obtaining the BVP signal \cite{dehaan2013robust,li2014remote,tasli2014remote,tulyakov2016self,wang2016algorithmic}. 
The blind source separation methods require an additional selection step to extract the most informative BVP signal.
To avoid this, we make use a `fixed' extraction scheme in our method.
%\jvg{Sort citations numerically, see my writing guidelines}. \jvg{Are the organizers cited? }
%\hl{Our proposed method falls under the category of `fixed' extraction scheme methods and thereby avoids the signal selection step.} 
%\mb{We can say multiple things as for why: We don't want an additional 'selection' step after seperating the signals. Based on findings of paper, that showed an improvement on ICA based methods for movement scenarios. Or because of our own experiments earlier where we found not so good performance}
%\hl{Our proposed method lists itself  in the latter partition.} 
%\jvg{any motivation for why?} \jvg{avoid reference words like 'latter'}

%\jvg{Can you help the reader to group this paragraph in her mind map? Add a sentence what it is about (see writing guidelines: 'related work') }
Among the `fixed' methods, multiple stand out and serve as inspiration and foundation for this work.
Tasli \etal~\cite{tasli2014remote} presented the first face modelling based signal extraction method and utilized detrending~\cite{tarvainen2002advanced} based filtering to estimate BVP and heart rate.
The CHROM~\cite{dehaan2013robust} method use a ratio of chrominance signals which are obtained from RGB channels followed by a skin-tone standardization step.
Li \etal~\cite{li2014remote} proposed an extra illumination rectification step using the colour of the background to counter illumination variations.
The SAMC~\cite{tulyakov2016self} method proposes an approach for BVP extraction in which regions of interest are dynamically chosen using self adaptive matrix completion. 
The Plane-orthogonal to skin (POS)~\cite{wang2016algorithmic} method improves on CHROM. 
It works by projecting RGB signals on a plane orthogonal to a normalized skin tone in normalized RGB space, and combines the resulting signals into a single signal containing the pulsatile information.
%The authors presenting the POS method mention that ``\dots neither detailed algorithm optimization nor dedicated signal processing were considered for attaining the highest accuracy[...]"\cite{wang2016algorithmic}. 
We take inspiration from Tasli \etal~\cite{tasli2014remote} and further build upon POS~\cite{wang2016algorithmic}. 
We introduce additional signal refinement steps for accurate peak detection to further improve HR and HRV analysis.
%\jvg{any motivation for why?}
%\mb{Could add that we use them because of their descent performance}

%\vspace{-0.2cm}
\paragraph{Deep learning based rPPG methods}
%\todo{Metion They use DL for skin pixel selection + filtering combined?; Drawback: slow + requires explicit training; }
% -Most recent trend has seen rise in Deep learning related techniques.
% -AFAWK only attempted three times up until now. \todo{Cite deepphys, spetlik, and the archive paper}
% -Use DL for ROI/ Skin-pixel selection: Slow requires specific training. 
% We overcome this by relying on AAM for skin pixel selection and our method works real-time.
Most recent works have applied deep learning (DL) to extract either heart rate or the BVP directly from camera images. They rely on the ability of deep networks to \textit{learn} which areas in the image correspond to heart rate. This way, no prior domain knowledge is incorporated and the system learns rPPG concepts from scratch.
%Only three works explore these possibilities.
DeepPhys~\cite{chen2018deepphys} is the first such end-to-end method to extract heart and breathing rate from videos.
HR-Net~\cite{vspetlikvisual} uses two successive convolutional neural networks~\cite{LeCun98Conv} to first extracts a BVP from a sequence of images and then estimate the heart rate from it.
Both show state-of-the-art performance on two public datasets and a number of private datasets.
%Most recently PhysNet \cite{yu2019recovering} was proposed which is using a deep spatiotemporal convolutional network to measure both HR and HRV features.% and claims to be the first end-to-end network that extracts a BVP.
Our presented algorithm makes use of an active appearance model \cite{vanKuilenburg2005model} to select regions of interest to extract a heart rate signal from. Due to this, no specific rPPG training is required while prior domain knowledge is more heavily replied upon.
%This allows our model to not depend on training, as deep learning related methods do. 

%In addition, none of the previous methods report execution times for their methods.
%Based on the architecture, it is unlikely that either of them is capable of running in real time.\todo{For Amogh: why do we think that}

\paragraph{HRV from PPG/rPPG}
% \todo{cite some pysc. papers about HRV itself; 
% Some non-RGB camera methods have attempted this (5-channel, etc); 
% very recently some methods have tried to get HRV from rPPG webcam (physnet + other paper(s));
%  } 
 % -Starting from Poh 2011 some papers have tried to get HRV using rPPG
% -Needs highly accurate measurements
% -(Explain shortly what HRV is)
% - Find out if someone tried this on the PURE dataset. 
% -
Some past methods have also attempted extracting heart rate variability from videos \cite{alghoul2017heart,poh2010advancements,sun2012noncontact}. 
%\jvg{sort citation numerically}.
A good overview is provided by Rodriguez \etal~\cite{rodriguez2018video}.
% \hl{
% To clarify: 
% "Heart rate is the number of heartbeats per minute. Heart rate variability (HRV) is the fluctuation in the time intervals between adjacent heartbeats.
% HRV indexes neurocardiac function and is generated by heart-brain interactions and dynamic non-linear autonomic nervous system (ANS) processes." \cite{shaffer2017overview}.
% }
Because of the way HRV is calculated, it is crucial that single beats are detected accurately with a high degree of accuracy.
Methods that otherwise show good performance in extracting HR can be unsuitable for HRV analysis, since they may not provide beat locations.
Rodriguez \etal~\cite{rodriguez2018video} evaluate their baseline rPPG method for HRV estimation. Their method is based on bandpass filtering the green channel from regions of interest. 
However, their results are only reported on their own dataset, which makes direct comparison difficult.
%\hl{which makes a comparison of their method to ours less accurate.} 
% \jvg{conclude why this is problematic }.
%Most recently \cite{yu2019recovering} reported their HRV performance on the publicly available MANHOB-HCI dataset using the earlier described PhysNet. 
Our method also estimates heart rate variability by obtaining precise temporal beat locations from the filtered BVP signal.
%\hl{In our method, we finely time heat beats by filtering the BVP signal to obtain a good HRV estimation.} 
%\mb{In our method we show promising starts for accurate peak detection and real time HRV estimation. }
%\ag{@Jan: Maybe we should not cite this paper. It's v. recent and only published on arxiv until now. Drawback of including them is they compute HRV from MAHNOB, while we dont. Reviewers might say we should do that too. We are not accurate enough on it.} \jvg{OK, I don't like to not cite openly. But OK, lets leave it out as long as we put it back in for our camera ready version :) } 

%% file: method.tex
\section{Method}
\label{sec:meth}

We present a method for extracting heart rate and heart rate variability from the face using only a consumer grade webcam.
%\jvg{not a fan of frameworks.. I like method}
Figure \ref{fig:overview} shows an overview of this method along with a summarized description.
% of it is as follows:
% an Active Appearance model~\cite{vanKuilenburg2005model} is used to detect regions of interest in the face as well as head orientation.
% Spatial averages of these regions are gathered over a time window into colour signals. %\jvg{what is that?}.
% These signals are resampled and projected onto a single signal based on POS~\cite{wang2016algorithmic}, which works by removing the influence of skin tone and specular reflections to obtain a cleaner BVP signal.
% %\hl{which obtains a cleaner BVP from RGB by removing influences of skin tone and specular reflections on these signal.}
% %\jvg{can you give a high level explanation of POS in a few words? A paper should be self-contained}.
% By tracking the head pose changes, noise caused by head movements is suppressed from the resulting signal.
% A wideband filtered version of this signal is then used to roughly estimate the current heart rate.
% %A wideband filtered version of the resulting signal is used in combination with head pose based motion noise suppression to roughly estimate the current heart rate.
% This rough estimate is in turn used to select a narrow band pass filter which is used to extract a clean BVP signal.
% Multiple windows of these BVP signals are overlap added and peak detection is performed on the resulting signal.
% Based on these peaks, heart rate and heart rate variability are calculated.
% %\jvg{Do we have space for a pipeline figure? Is that Fig 1?}

\subsection{Skin pixel selection}
\label{sec:ROI}
The first step in the pipeline includes face finding \cite{viola2001rapid} and fitting an active appearance model (AAM)~\cite{vanKuilenburg2005model}. This AAM is then used to determine facial landmarks and head orientation. 
The landmarks are used to define a region of interest (RoI) which only contains pixels on the face belonging to skin. This allows us to robustly track the pixels in this RoI over the course of the whole video.
Our RoI consists of the upper region of the face excluding the eyes. An example of this can be seen in Figure \ref{fig:overview} and \ref{fig:screenshots}.
The head orientation is used to measure and track the pitch, roll, and yaw angles of the head per frame.
Across all pixels in the RoI, the averages for each colour channel (R,G,B) is computed and tracked (concatenated) to create three colour signals.

\subsection{Signal extraction}
\label{sec:Signal}

The colour signals and the head orientation angles are tracked over a running window of 8.53~seconds. This window duration corresponds to 256~frames at 30~fps, or 512~frames at 60~fps.
%If a face was lost shorter than half the size of the running window, the trace up to this point was mirrored for the amount of time the face was lost.
All signals are resampled using linear interpolation to counteract variations in frame rates of the input. They are resampled to either 30 or 60~fps, whichever is closer to the frame rate of the source video.
Subsequently, the three colour signals from R, G and B channels are combined into a single rPPG signal using the POS method~\cite{wang2016algorithmic}.
The POS method filters out intensity variations by projecting the R, G and B signals on a plane orthogonal to an 
experimentally determined normalized skin tone vector. 
%\mb{From Tom: what is experimentally determined skin colour...? would be nice if you can say that this is done for each subject (that would at least clarify it a bit)}
The resulting 2-D signal is combined into a 1-D signal with one of the input signal dimensions being weighted by an alpha parameter that is the quotient of the standard deviations of each signal. 
This ensures that the resulting rPPG signal contains the maximum amount of pulsating component.
%This results in the raw rPPG trace.

%\vspace{-15pt}
\subsection{Signal filtering}
\label{sec:filtration}
\paragraph{Rhythmic motion noise suppression}
A copy of the extracted rPPG signal as well as the head-orientation signals are converted to frequency domain using Fast Fourier Transform.
% \jvg{Capital FD? Fourier is God? ;) } 
The three resulting head-orientation spectra (one each of pitch, roll, and yaw) are combined into one via averaging.
%\jvg{OK, no need to mention this answer here, but it does make me wonder why not use a 3D fourier transform on all 3 head orientation channels?} \mb{huh?}.
This is then subtracted from the raw rPPG spectrum after amplitude normalization. 
%\mb{Tom asked: what kind of normalization, but I think we left in intentionally out}
This way, the frequency components having a high value in the head-orientation spectrum are attenuated in the rPPG spectrum.
%To remove frequency components stemming from periodic movements of the head (motion noise), the resulting average head-orientation spectrum is subtracted from the rPPG spectrum. 
Subsequently, the frequencies outside of the human heart rate range (0.7 - 4~Hz / 42 - 200~bpm) are removed from the spectra.
%All rPPG spectra were averaged over the last window time, or less during the first seconds of the video.

\paragraph{Wide \& narrow band filtering}
The highest frequency component inside the resulting spectrum is then used to determine the passband range of a narrow-bandpass filter with a bandwidth of 0.47~Hz.
Such a filter can either be realized via inverse FFT or a high order FIR filter (e.g. \texttildelow50\textsuperscript{th} order Butterworth).
%The highest frequency component inside the resulting spectrum is then used to select one of 13 50th order Butterworth Band pass filters spanning the range of 0.7 - 4 Hz.
%Each filter had a bandwidth of 0.47Hz and approximately 50\% overlap to the next filter.
The selected filter is then applied to the original extracted rPPG signal to produce noise-free BVP. 
%\todo{other word for purified/clean?}
%\jvg{Other word for 'clean'? }

\subsection{Post processing}
\label{sec:postprocess}
%During training, the gradients computed from the loss layer reach the CAM layer through the global pooling layer. 
%The connecting weights between the CAM and the previous conv layers are updated based on the distribution/flow of the gradients defined by the type of global pooling layer used.
%Hence, the choice of global pooling layer and its distribution of gradients to bottom layers is an important consideration for this framework for weak supervision. %of weakly supervised object localization.
%To generate the most confident version of the signal, overlap adding \jvg{unclear sentence} is applied similar to how it is done in \cite{dehaan2013robust,dehaan2014improved,wang2016algorithmic} \jvg{sort}.
To prevent minor shifts in the locations of the crest of each beat over multiple overlapping running windows, the signals from each window are overlap added with earlier signals~\cite{dehaan2013robust,dehaan2014improved,wang2016algorithmic}.
First, the filtered rPPG signal is normalized by subtracting its mean and dividing it by its standard deviation.
During resampling of the signal, the number of samples to shift is determined based on the source and resampled frame rates.
The signal is then shifted back in time accordingly and added to the previous/already overlapped signals.
Older values are divided by the times they have been overlap added, to ensure all temporal locations lie in the same amplitude range.
% to prevent them from increasing in size.
Over time, a cleaner rPPG signal is obtained from this.
%weighted down according to

\subsection{Output calculation}
\label{sec:hrandhrv}
%\jvg{avoid having 'tion' on a separate line}
% \begin{figure}
% 	\floatbox[{\capbeside\thisfloatsetup{capbesideposition={left,center},capbesidewidth=0.45\textwidth}}]{figure}[\FBwidth]
% 	{\caption{Example of heart rate variability computation: Even when the heart rate (HR) is almost constant, the underlying inter-beat intervals (IBIs) can have many fluctuations. This is detected by rising squared successive differences (SSD), a measure of heart rate variability.}
% 	\label{fig:HRVexample}}
% 	{\includegraphics[width=0.5\textwidth]{HRV-example.pdf}}
% \end{figure}

\begin{figure}
	\centering
	\fbox{\includegraphics[width=0.6\textwidth]{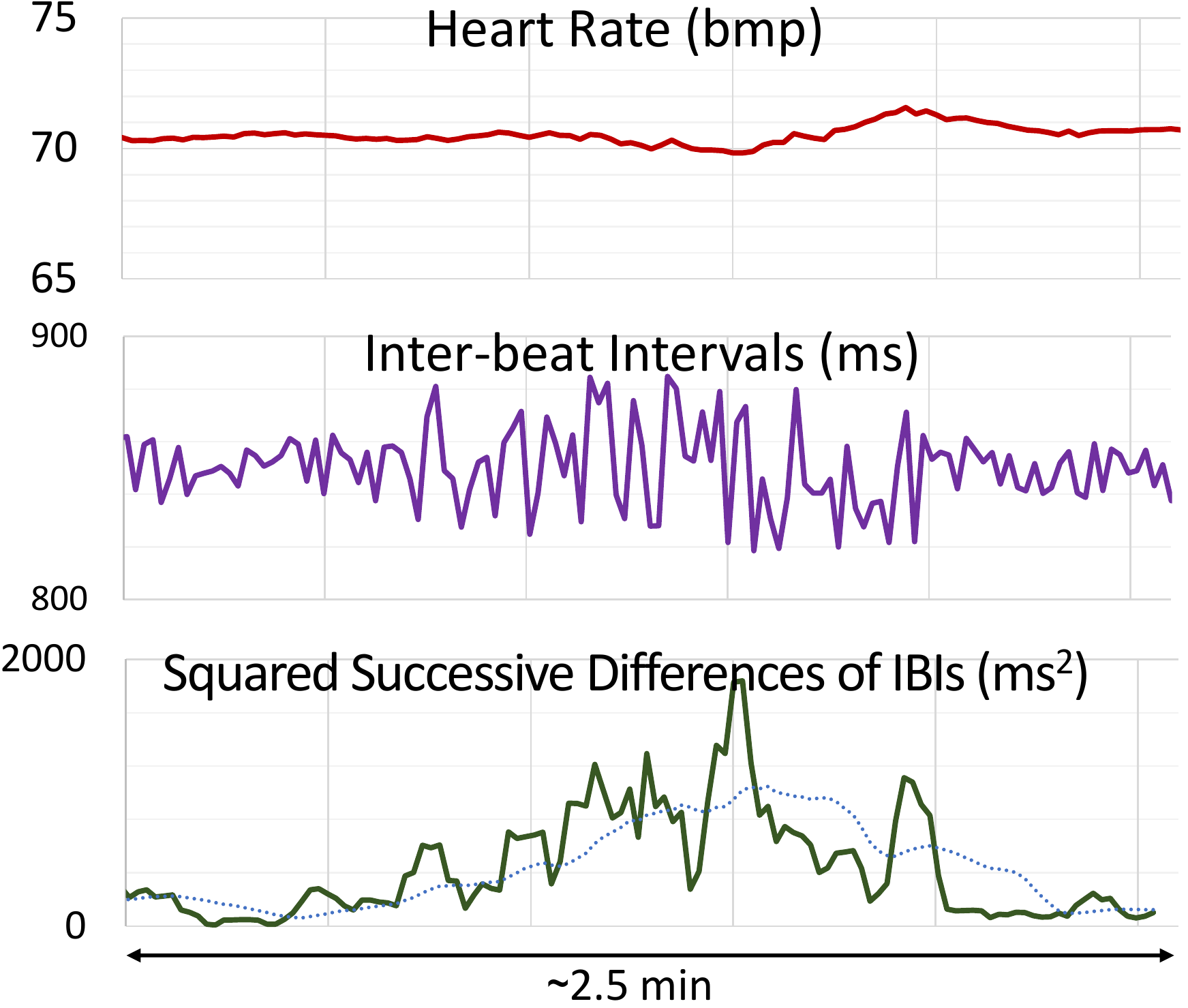}}
	\caption{Example of heart rate variability computation: Even when the heart rate (HR) is almost constant, the underlying inter-beat intervals (IBIs) can have many fluctuations. This is detected by rising squared successive differences (SSD), a measure of heart rate variability (HRV).
	\ifsqueeze\vspace{-0.3cm}\fi}
	%\mb{Should we mention SDs are squared?
	%\todo{This figure is never referred!}
	%\jvg{'a lot of' = 'many'}
	\label{fig:HRVexample}
\end{figure}

Once a clean rPPG signal is obtained, we can perform peak detection on it to locate the individual beats in time in the signal.
From the located beats, heart rate and heart rate variability can be calculated.
To do this, we first extract the inter-beat-intervals (IBIs) from the signal, which are the time intervals between consecutive beats.

\paragraph{Heart rate calculation}
Heart rate is calculated by averaging all IBIs over a time window, and computing the inverse of it. That is, 
$\text{HR}_w = 1/{\overline{\text{IBI}}_w}$,
% \begin{equation}
% 	\scriptstyle
% 	%\text{HR}_w = 1/\mean(\text{IBI}_i),
% 	\text{HR}_w = 1/{\overline{\text{IBI}}_w},
% 	%\text{HR}_w = \frac{60}{t_b - t_e} \sum_{i=t_b}^{t_e} \text{IBI}_i,
% 	%\text{where } \overline{\text{IBI}}_w \text{is the mean of all inter-beat intervals that fall withing the time window} w.
% \end{equation}
where $\overline{\text{IBI}}_w$ is the mean of all inter-beat intervals that fall within the time window $w$.
%where $\text{IBI}_i$ represents the $i$\textsuperscript{th} inter-beat interval, $t_b$ represents the start time of the window, $t_e$ represents the end time of the window and $w$ represents window spanned by $t_b$ and $t_e$.
The choice of this time window can be based on the user's requirement (e.g. instantaneous HR, long-term HR).

\paragraph{Heart rate variability calculation}
Multiple metrics can be computed to express the measure of heart rate variability in different units.
%To compute heart rate variability, multiple calculation methods exist to compute different metrics. 
In this work, we focus on one of the most popular time-domain metric for summarizing HRV called the root mean square of successive differences (RMSSD)~\cite{huang2016measurement,malik1996heart,rodriguez2018video}, expressed in units of time.
As the name suggests, this is computed by calculating the root mean square of time difference between adjacent IBIs:

%\jvg{use backslash text for RMSSD:}
\begin{equation}
	\scriptstyle
    \text{RMSSD} = \sqrt{\frac{1}{N-1} \Big(\sum_{i=0}^{N-1} (\text{IBI}_i - \text{IBI}_{i+1}) ^ 2 \Big)},
\end{equation}
where $\text{IBI}_i$ represents the $i$\textsuperscript{th} inter-beat interval, and $N$ represents the number of IBIs in the sequence. A graphical example of such HRV calculation is shown in Figure \ref{fig:HRVexample}.

In addition, we also compute two frequency-domain metrics of HRV, simply known as Low-frequency (LF) and High-frequency (HF) band~\cite{malik1996heart} (as well as a ratio of them), that are commonly used in rPPG HRV literature \cite{alghoul2017heart,mcduff2014improvements,rodriguez2018video}.
%\jvg{Why? Motivate.}
%\mb{Not sure if they state it explicitly but from the papers they show it becomes clear}
%\jvg{add spaces before (using) brackets}.
The LF and HF components are calculated using Welch's power spectral density estimation.
Since Welch's method expects evenly sampled data, the IBIs are interpolated at a frequency of 2.5Hz and zero padded to the nearest power of two.
The power of each band is calculated as total power in a region of the periodogram: the LF band from [0.04 to 0.15~Hz], and the HF band from [0.15 to 0.4~Hz].
Details about these metrics can be found in \cite{shaffer2017overview}.
Both metrics are converted to normalized units by dividing them by the sum of LF and HF.
%\hl{To make comparison with earlier works eaiser both metrics are converted to normalized units by dividing LF and HF respectively by the sum of LF and HF. }
%\jvg{yes, and add a citation for them} 
%\jvg{why can they be used at all? If they are in the Fourier domain they need to be completely regular, no?}

%% file: experiments.tex
\section{Experiments and Results}
\label{sec:exp}

\begin{figure}
	\centering
	\includegraphics[width=\textwidth, trim={0 3.82cm 2cm 0}, clip]{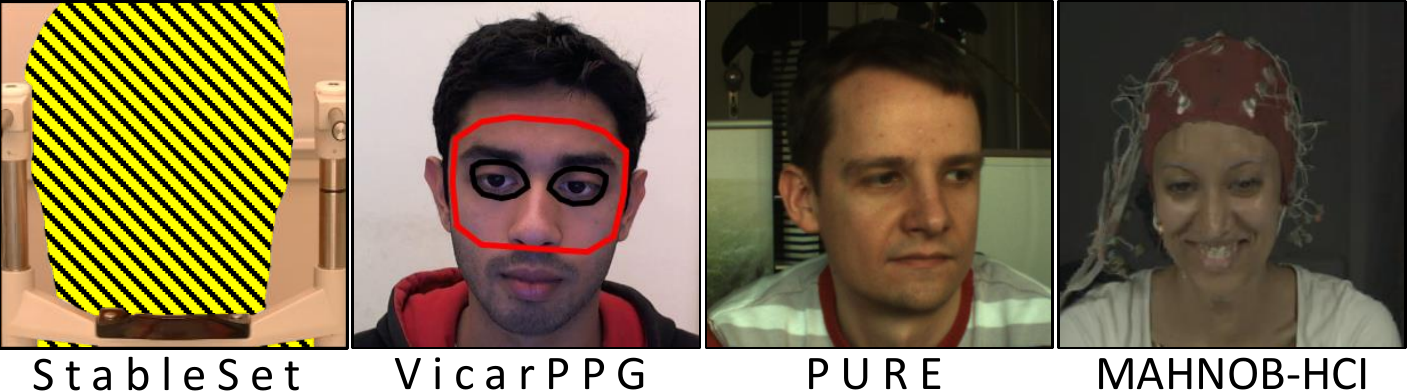}
	\caption{Examples of images from (left to right) the StableSet, VicarPPG, PURE, and MAHNOB-HCI datasets. The example from VicarPPG dataset shows the RoI overlaid on the face. The subjects in StableSet were physically stabilized using the shown chin rest (face removed for privacy reasons). The subjects in PURE perform deliberate head movements. The videos from MAHNOB-HCI suffer from high compressions noise.
	%\todo{get thicker RoI line.}
	}
	\label{fig:screenshots}
	%\vspace{-8pt}
\end{figure}

%\todo{Think of more figures!}
%\ag{@Jan: We still need to definitely decide if we want to include our results on the MAHNOB-HCI dataset.} 
%\jvg{I'm OK to take it out if you would prefer that}
\subsection{Datasets}
Some example frames from the datasets used in this paper can be seen in Figure~\ref{fig:screenshots}.
%\jvg{Remove whole sentence, nothing is added by 'We perform our experiments on following rPPG video datasets that come along with PPG or ECG ground truth signals.'}

\paragraph{StableSet rPPG dataset}
To make a proof-of-concept test of our proposed rPPG method, we recorded the StableSet rPPG dataset\ificcvfinal\footnote{As part of research work at the Human-Technology interaction group, Eindhoven University of Technology.
We acknowledge and thank dr. Daniel Lakens and the research team for their valuable contributions.}\fi.
This video dataset consists of 24 subjects recorded at 25~fps in 1920$\times$1080 resolution using a RGB camera and a 1~KHz medical-grade electrocardiogram (ECG) device connected to the subjects. %Daniël
The subjects' head movements were physically stabilised using a chin rest with the intention of minimizing motion induced noise in rPPG measurements.
The subjects were recorded while watching emotion inducing video stimuli as well as playing the game of Tetris at varying difficulty levels. This was done with the intention of inducing HRV changes. 
%\mb{We can give the accurate names of the camera and ECG for further information}
%\todo{show screenshot?} \jvg{If you have that, and if there is space, yes :) }

\paragraph{VicarVision rPPG dataset - VicarPPG}
The VicarPPG dataset~\cite{tasli2014remote} consists of 20 video recordings of 10 unrestrained subjects sitting in front of a webcam (Logitech c920).
The subjects were recorded under two conditions: at rest while exhibiting stable heart rates, and under a post-workout condition while exhibiting higher heart rates gradually reducing.
The videos were originally recorded at 1280$\times$720 resolution with an uneven variable frame rate ranging from as low as \texttildelow5~fps up to 30~fps.
The frames were later upsampled and interpolated to a fixed 30~fps frame rate video file.
The ground truth was obtained via a finger pulse oximeter device (CMS50E).

\paragraph{Pulse Rate Detection Dataset - PURE}
The PURE dataset~\cite{stricker2014non} comprises of 60 videos of 10 subjects.
Every subject is recorded under 6 different conditions with increasing degree of head movements including talking, slow and fast translation, small and large rotation.
The videos were recoded at 30~fps in 640$\times$480 resolution with no compression (lossless), and the ground truth was obtained via a pulse oximeter (CMS50E).
% \jvg{next sentence repeats this; merge sentences}.

\paragraph{MAHNOB-HCI Tagging rPPG Dataset}
This dataset consists of 527 videos of 27 subjects, along with 256~Hz ECG ground truth recording. 
The videos were recoded at 61~fps in 780$\times$580 resolution and compressed to a high degree. 
To make our results comparable to previous work, we extract the same 30 second video duration from these videos (frames [306 - 2135]) and only analyse these.
%\jvg{dont mention what you dont use. Mentioning this larger dataset only makes it more confusing and it is not even  relevant} 

\subsection{Heart Rate Analysis}

\begin{figure}
	\centering
	\subfloat[][StableSet]{\includegraphics[width=.49\textwidth, trim={0 0 0 0.79cm}, clip]{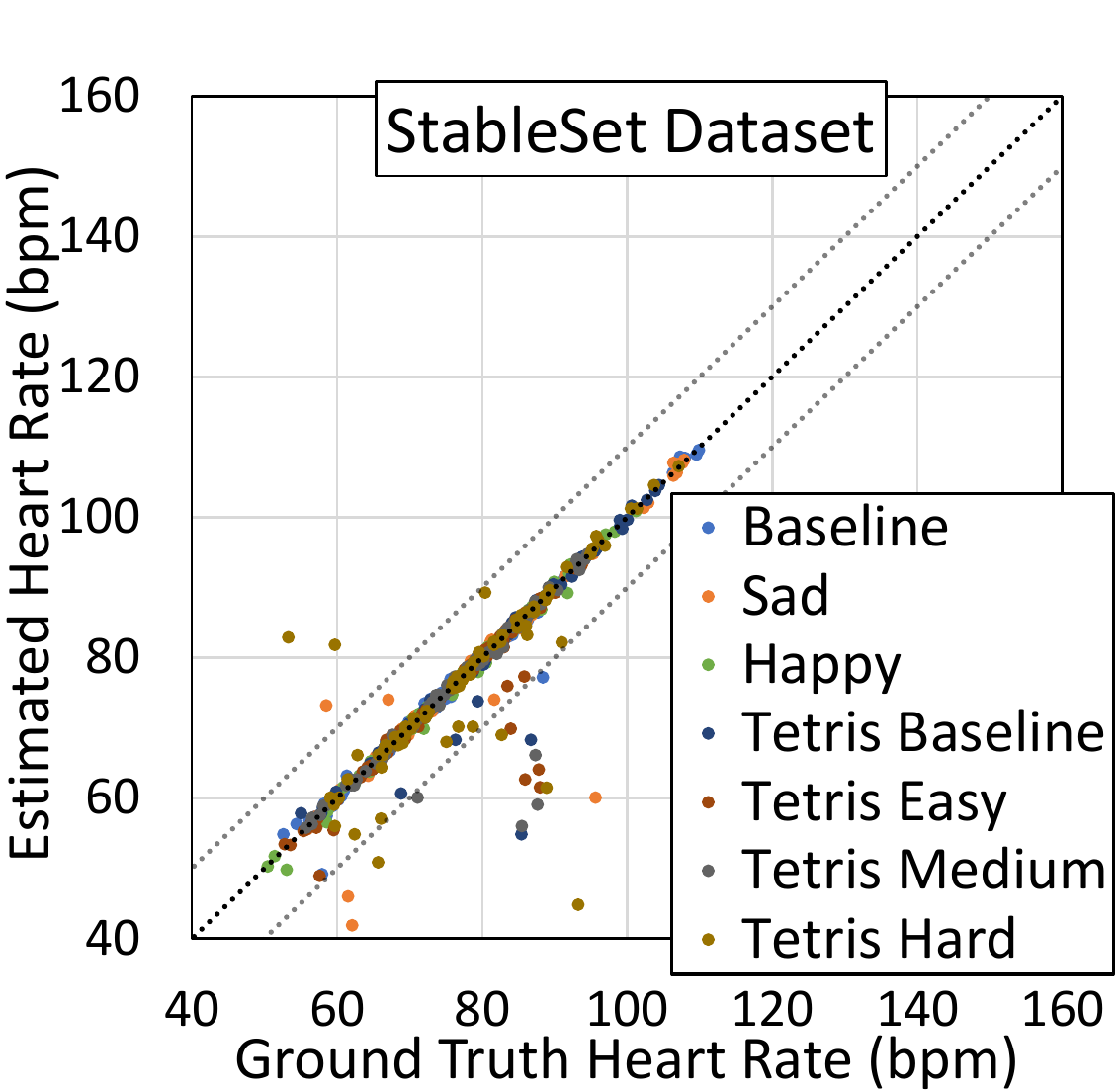}\label{fig:HRresults-StableSet}} 
	\subfloat[][VicarPPG]{\includegraphics[width=.49\textwidth, trim={0 0 0 0.79cm}, clip]{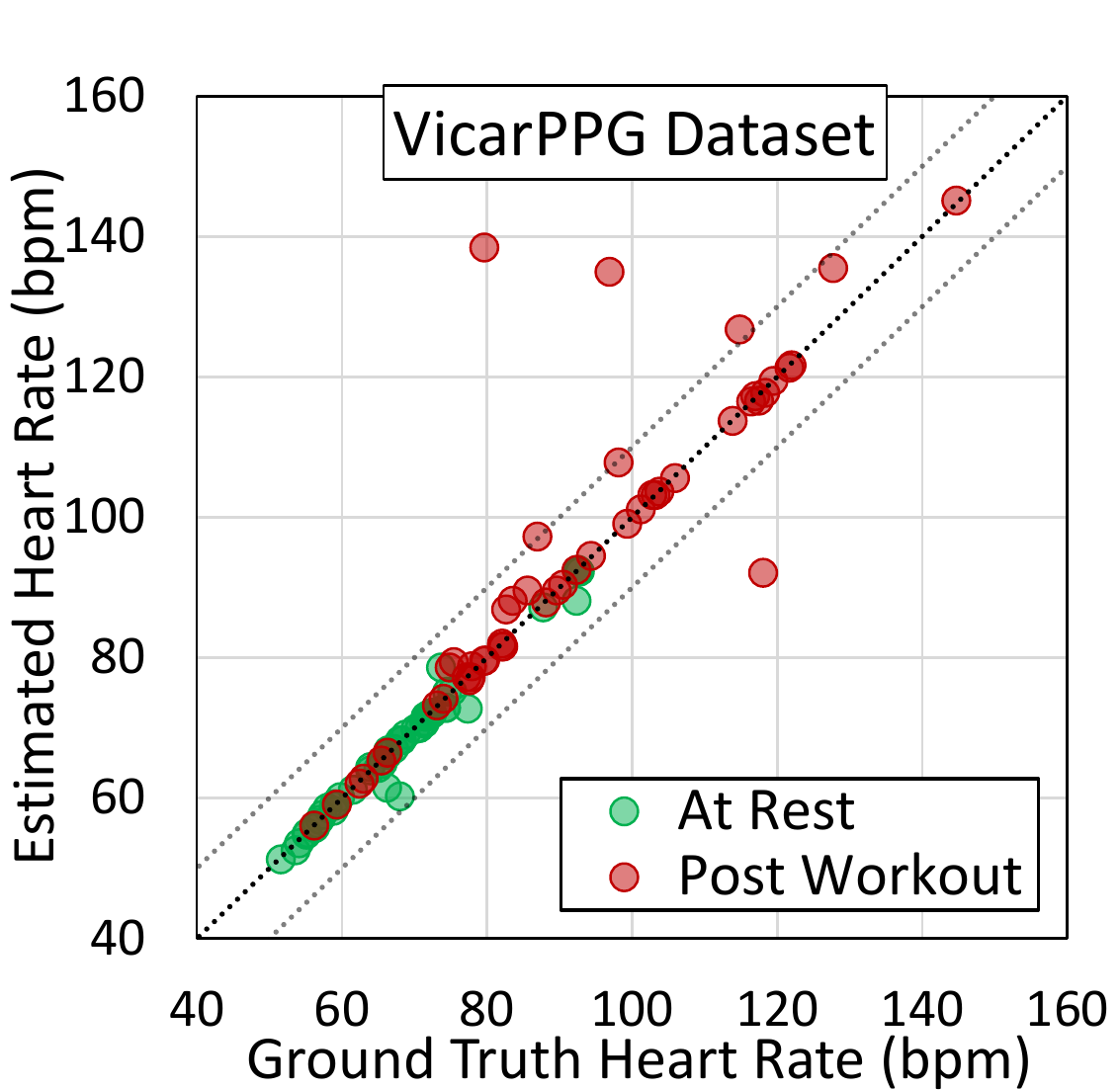}\label{fig:HRresults-VicarPPG}}\\
	\subfloat[][PURE]{\includegraphics[width=.49\textwidth, trim={0 0 0 0.79cm}, clip]{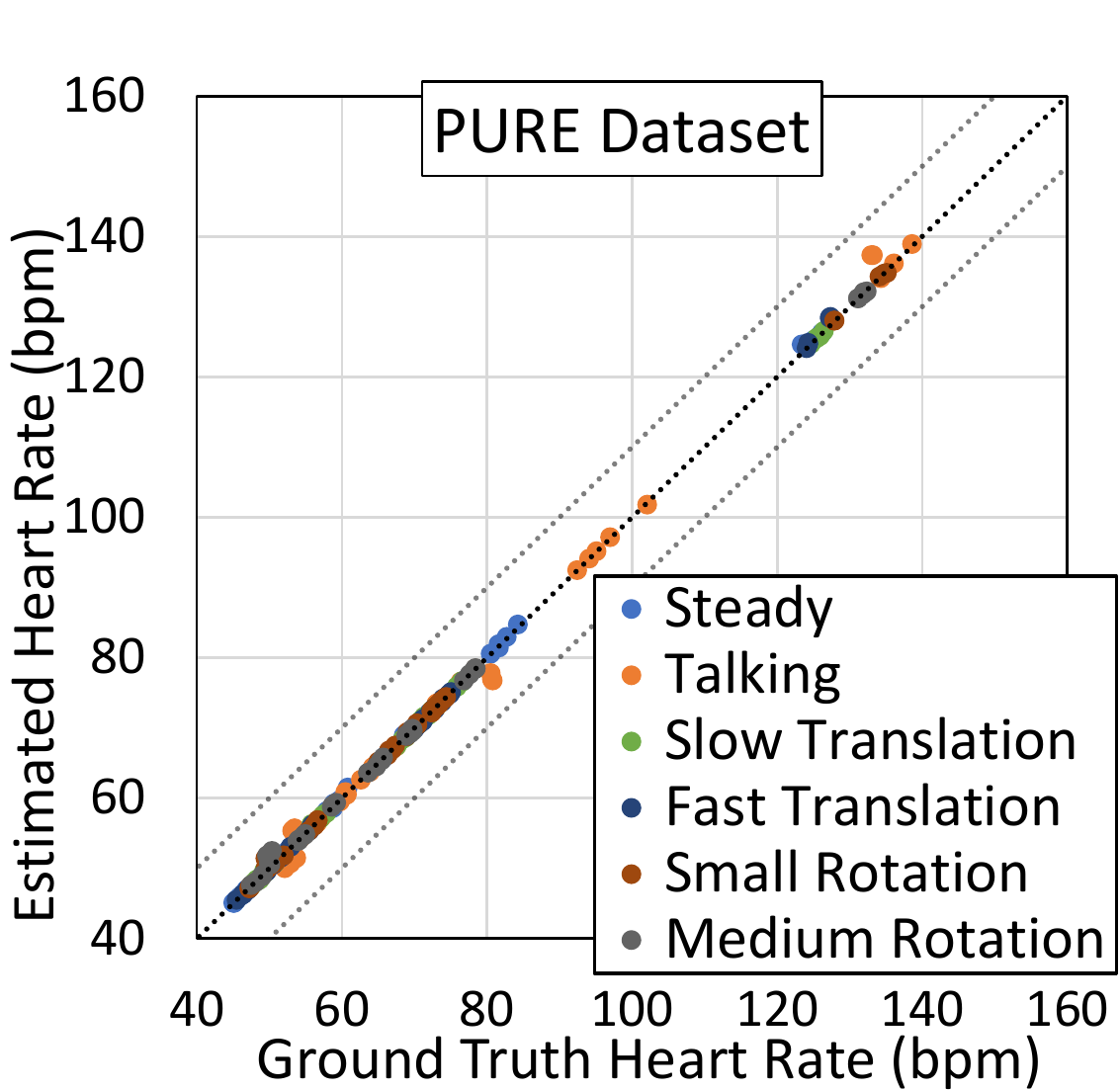}\label{fig:HRresults-PURE}} 
	\subfloat[][MAHNOB-HCI]{\includegraphics[width=.49\textwidth, trim={0 0 0 0.79cm}, clip]{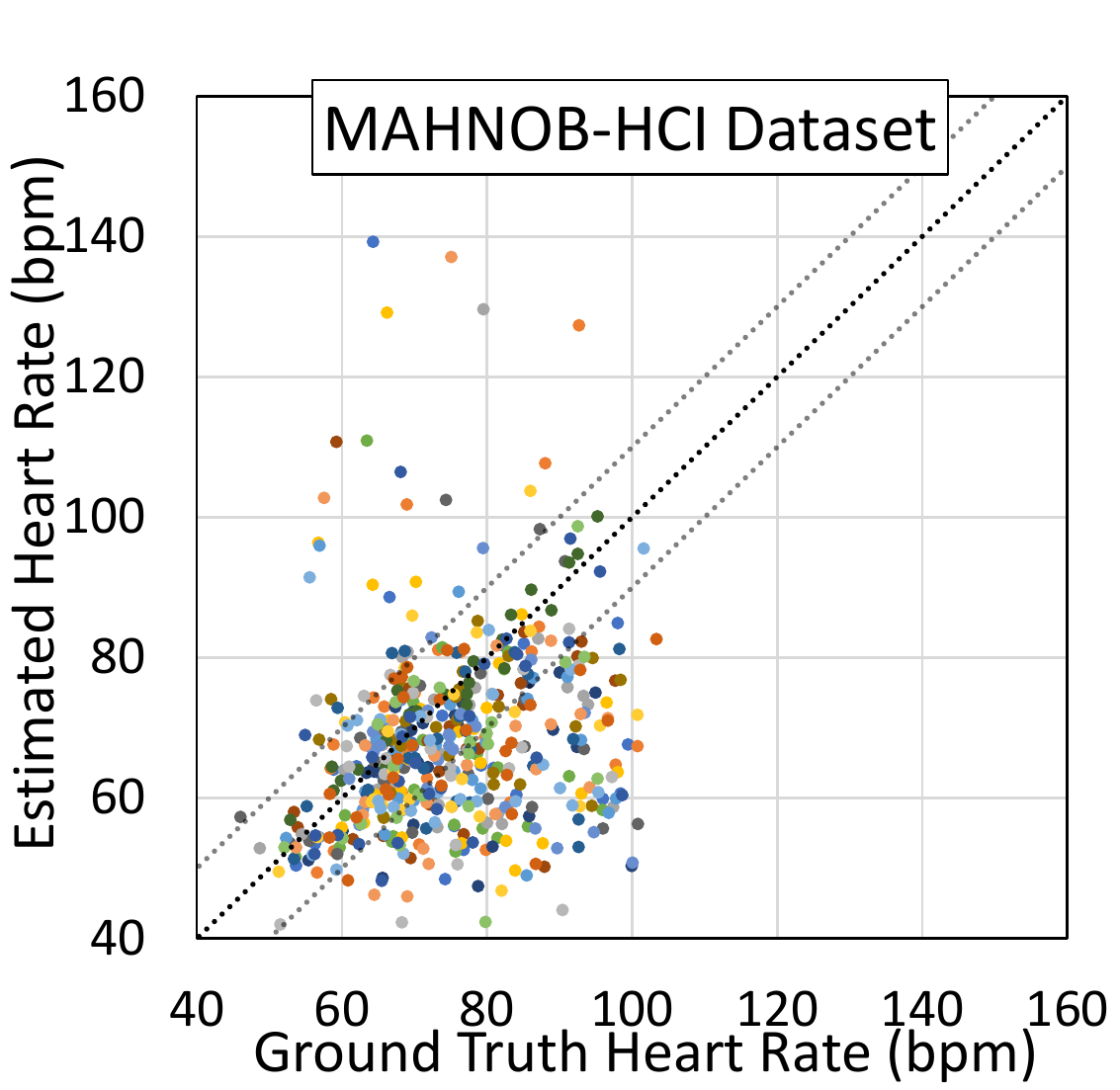}\label{fig:HRresults-MAHNOB}}
	\caption{Scatter plot of the predicted vs ground truth heart rates. Each point represents one video segment in the dataset: (a) StableSet (8.5s segment); (b) VicarPPG  (15s segments); (c) PURE (30s segments); (d) MAHNOB-HCI (30s segments).
	While high correlation between the ground truth and estimated heart rates can be seen in the first three datasets, the results on MAHNOB-HCI is worse. This can be attributed to its high compression noise.
	\ifsqueeze\vspace{-0.3cm}\fi}
	%\mb{Make text smaller? Gets confusion when reading through paper}

	%\jvg{Conclusion missing. what do you want me to see? A reader first goes in comic book mode: add a conclusion to each caption} 
	 %\jvg{Too small} 
	\label{fig:HRresults}
 \end{figure}

% \begin{figure}
% 	\centering
% 	\fbox{\includegraphics[width=\textwidth]{HRresults-plot-dummy.jpg}}
% 	\caption{Scatter plots of HR performance on all dataset?}
% 	\label{fig:HRresults}
% 	%\vspace{-8pt}
% \end{figure}

\begin{figure}
	\centering
	\includegraphics[width=\textwidth]{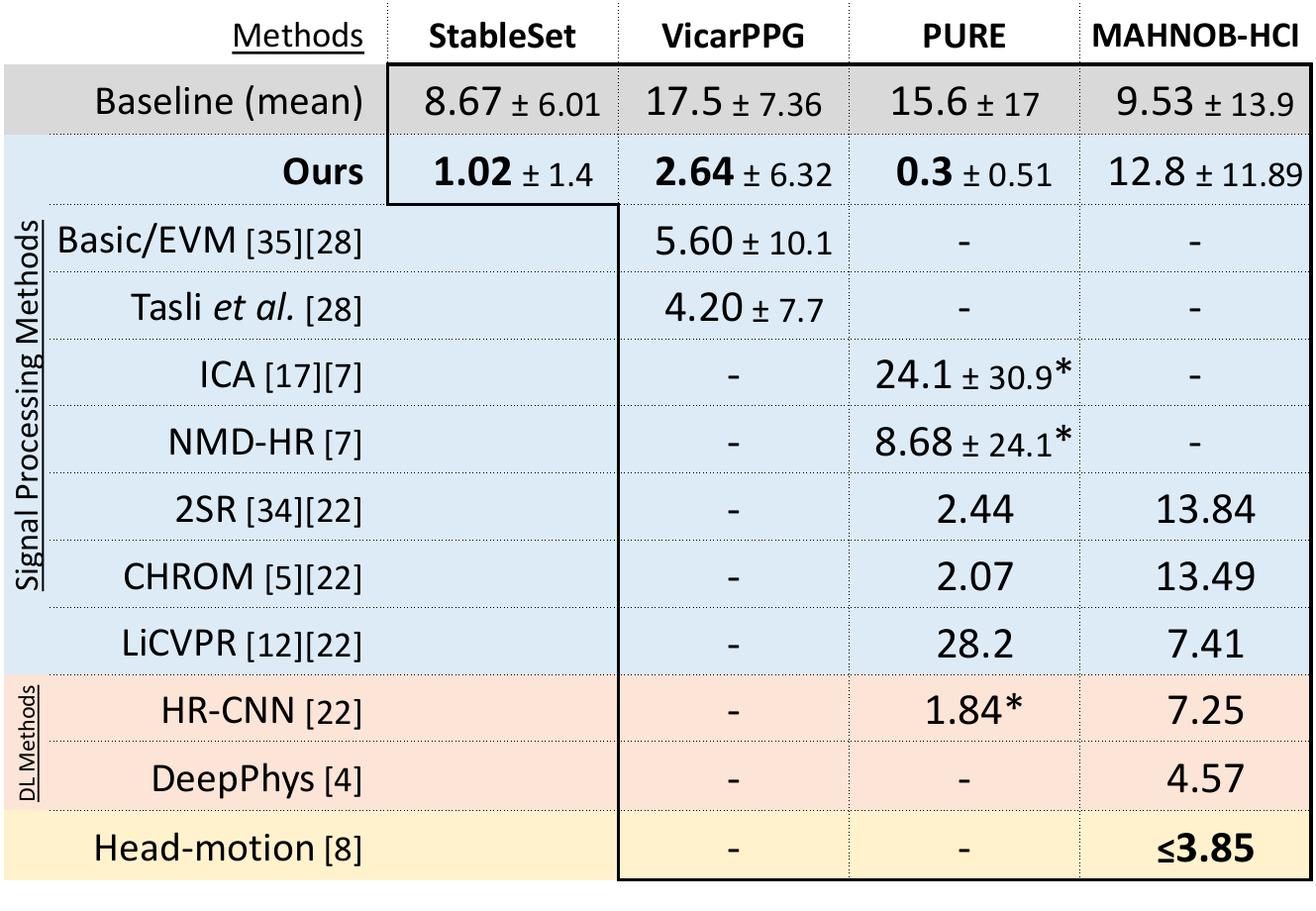}
	\captionof{table}{A comparison of the performances of various methods in terms the mean absolute error in beats per minute (bpm).
	Baseline (mean) represents the accuracy obtained by only predicting the mean heart rate of the dataset.
	*~represent accuracies obtained on a slight variations of the full dataset: ICA~\cite{poh2010non}\cite{demirezen2018remote}, NMD-HR~\cite{demirezen2018remote}, HR-CNN~\cite{vspetlikvisual} were tested on a 8, 8, 4-person subset of the PURE dataset respectively.
	$\scriptstyle\le$~represents root mean squared error, which is always greater than or equal to mean absolute error.
	The different colours separate the different categories of methods: signal processing, deep learning and head-motion based methods.
	Our proposed method obtains a high accuracy on StableSet, VicarPPG and PURE datasets and outperforms all previous methods.
	Accuracy on MAHNOB-HCI dataset is low similar to most other signal-processing methods, likely due to its high compression noise.
	Note that the simple mean predicting baseline obtains an accuracy quite close to most methods on this dataset.
	%\todo{EVM~\cite{Wu12Eulerian}\cite{tasli2014remote}, Tasli~\etal~\cite{tasli2014remote}. ICA~\cite{poh2010non}~\cite{demirezen2018remote}, NMD-HR~\cite{demirezen2018remote}, 2SR~\cite{wang2015novel}\cite{vspetlikvisual}, CHROM~\cite{dehaan2013robust}\cite{vspetlikvisual}, LiCVPR~\cite{li2014remote}\cite{vspetlikvisual}, HR-CNN~\cite{vspetlikvisual}, DeepPhys~\cite{chen2018deepphys}, Head-Motion~\cite{haque2016}.}
	%\todo{refine?}
	%\todo{cite papers mentioned here}
	%\todo{explain colours?}
	%\todo{explain stars *}
	\ifsqueeze\vspace{-0.3cm}\fi}
	%\jvg{Explain what colors mean} \jvg{Conclusion missing. what do you want me to see? A reader first goes in comic book mode: add a conclusion to each caption}
	\label{tab:HRresults}
	%\vspace{-8pt}
\end{figure}

\begin{figure*}
	\centering
	\subfloat{\includegraphics[width=.13\textwidth, trim={0 1.55cm 3cm 0}, clip]{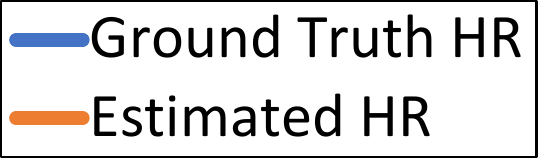}\label{HRexample-legend2}}\quad
	\subfloat{\includegraphics[width=.21\textwidth]{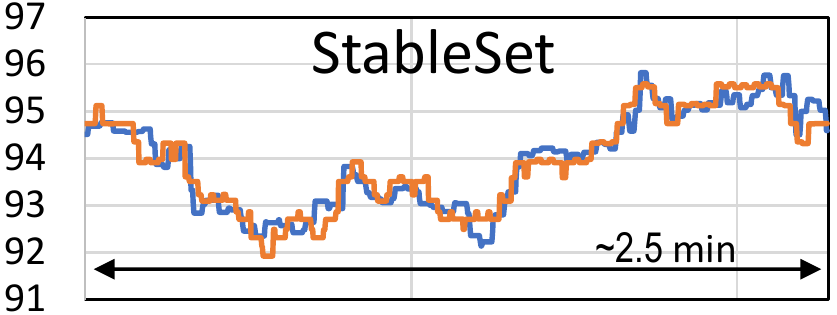}\label{HRexample1-StableSet}}
	\subfloat{\includegraphics[width=.21\textwidth]{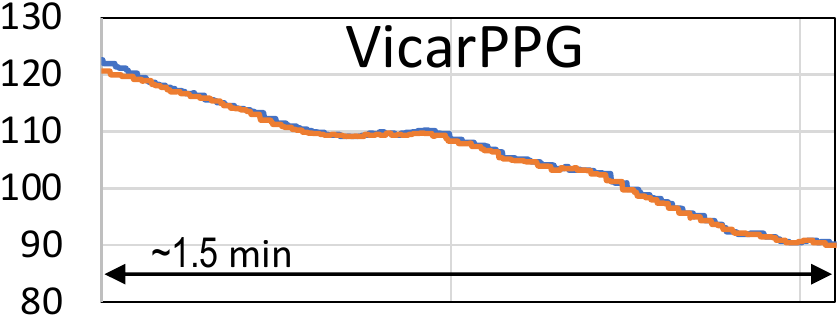}\label{HRexample1-VicarPPG}}
	\subfloat{\includegraphics[width=.21\textwidth]{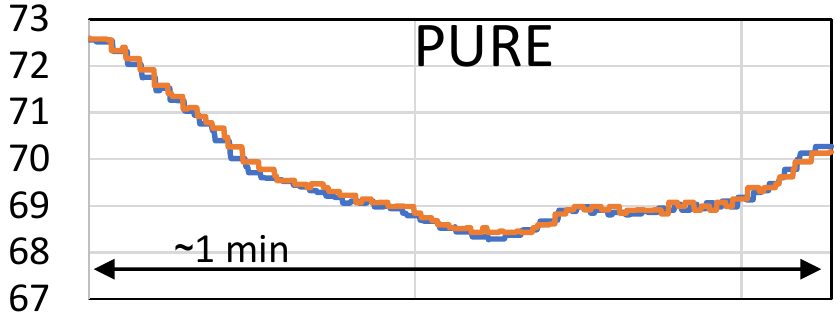}\label{HRexample1-PURE}}
	\subfloat{\includegraphics[width=.21\textwidth]{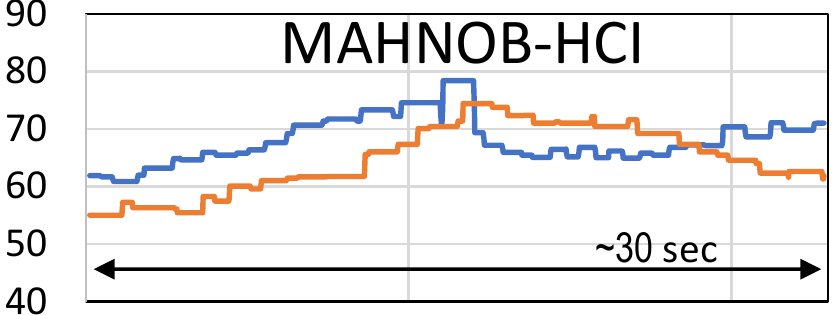}\label{HRexample1-MAHNOB}}\\
	\subfloat{\includegraphics[width=.21\textwidth]{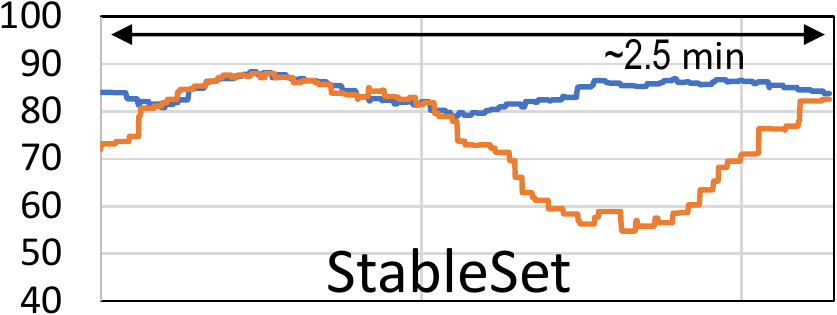}\label{HRexample2-StableSet}}
	\subfloat{\includegraphics[width=.21\textwidth]{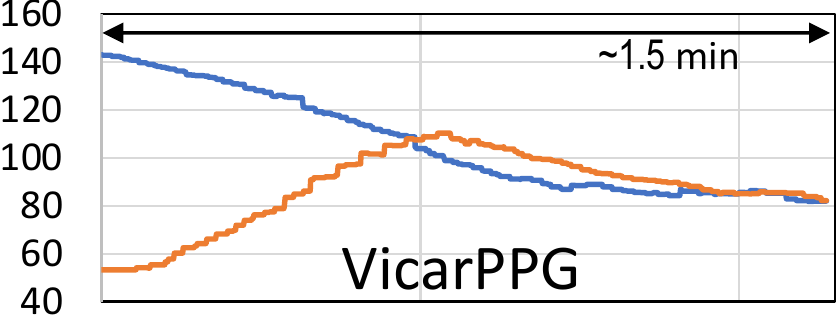}\label{HRexample2-VicarPPG}}
	\subfloat{\includegraphics[width=.21\textwidth]{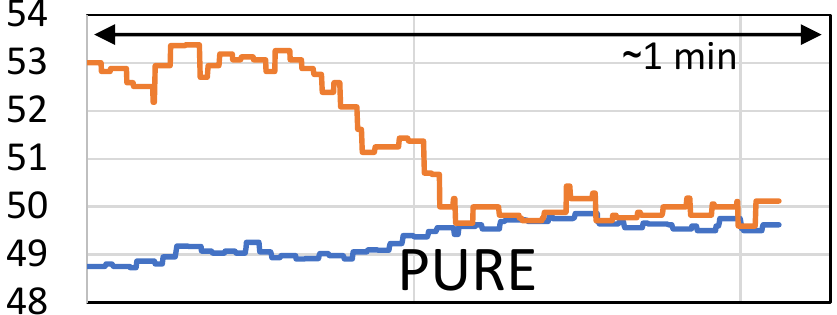}\label{HRexample2-PURE}}
	\subfloat{\includegraphics[width=.21\textwidth]{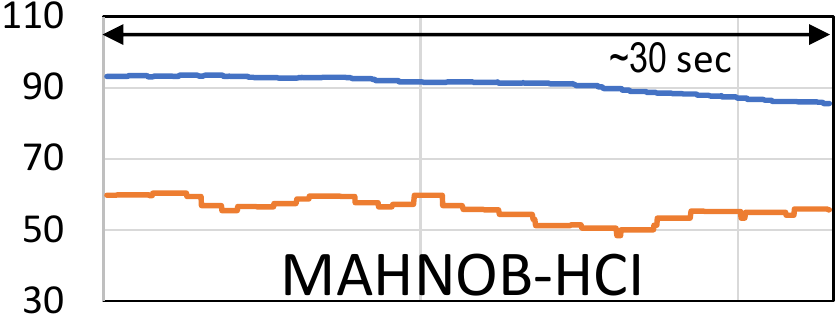}\label{HRexample2-MAHNOB}}\quad
	\subfloat{\includegraphics[width=.13\textwidth, trim={0 1.55cm 3cm 0}, clip]{HRexample-legend.pdf}\label{HRexample-legend}}
	%angle=90, origin=c,
	\caption{Examples of heart rate estimation from all datasets (x-axis --- time, y-axis --- heart rate; top row --- good examples, bottom row --- bad examples). When conditions are right, the estimated heart rate is able to follow the ground truth closely.
	The rare errors in StableSet and PURE are due to incorrect face modelling caused by occlusion (chin-rest) and deformation (talking) respectively.
	Prediction errors in VicarPPG are mostly in the high HR range due to low frame rate artefacts, while on MAHNOB-HCI they are due to compressions noise.
	\ifsqueeze\vspace{-0.3cm}\fi}
	%\jvg{too small. Suggestions: Make lines thicker? Place legend in fig? Maybe try using tabular instead of subfloats? } 
	%\jvg{Conclusion missing. what do you want me to see? A reader first goes in comic book mode: add a conclusion to each caption} 
	\label{fig:HRexamples}
 \end{figure*}

%\jvg{Read my writing guidelines about 'in order'} 
To assess the heart rate estimation accuracy of our method, we measure the deviation of the predicted heart rates from the ground truth in all the datasets. 
We express this deviation in terms of the mean absolute error (MAE) metric in beats per minute (bpm). 
This metric is the average of the absolute differences between predicted and true average heart rates obtained within a set time window. 
To make a fair comparison with other work, we set different time-window sizes for MAE computation per dataset to match the ones used in prior work: 8.5~secs, 15~secs, 30~secs, and 30~secs on StableSet, VicarPPG, PURE, and MAHNOB-HCI respectively. 
It should be noted that while shorter time-windows require more precise estimation, the choice of this window size did not affect our results significantly.
%It should be noted that while the choice of this window size did not affect our results significantly.

The summarised results of this heart rate analysis in comparison with other work can be in Table~\ref{tab:HRresults} and Figure~\ref{fig:HRresults}, while some qualitative examples can be seen in Figure~\ref{fig:HRexamples}.
On the StableSet, our proposed method obtains a low error rate of 1.02~bpm.
%\hl{This was obtained in spite of setting a small 8.5~secs averaging window, which means that on average the predicted heart rate over every 8.5~sec segment of the video was off by only around one beat per minute.}
%\mb{I think this contradicts what is said earlier, that the window does not have a large influence}
This high accuracy can be attributed to the fact that subjects' movements were physically stabilized via a chin-rest (see Figure~\ref{fig:screenshots}).
%\jvg{Read my writing guidelines on 'very'} 

On the VicarPPG and the PURE datasets, our method outperforms all previous methods by a large margin.
%Although the error rate is larger than that on the StableSet, it can still be considered quite low.
This is in spite of the subjects being unrestrained, exhibiting a wide range of changing heart rates, and performing a variety of large and small head movements.
The very high accuracy of 0.3~bpm on the PURE dataset can be attributed to the fact that the videos were lossless encoded and had no compression noise. All the noise was caused was purely due to head movements, and this was the main failure point of other methods. Our method is able to filter out this motion noise significantly well.
Conversely, on closer analysis, we found that the relatively lower average error on the VicarPPG dataset was primarily due to some segments in the videos having very low effective frame rates (\texttildelow5~fps). 
%\mb{original}
%Such drops can be seen in Figure \ref{fig:HRexamples}. 
This low frame rate approaches the Nyquist frequency for human heart rate analysis, which is a theoretical limitation that says sampling frequency must at least be twice the highest frequency to be measured. 
%\mb{should we elaborate?}
If videos severely suffering from such low frame rate artefacts are excluded (namely \texttt{06\_WORKOUT}, \texttt{08\_WORKOUT}), the error rate drops to \textbf{0.84{\raisebox{.2ex}{$\scriptstyle\pm$}}0.75~bpm}.
%\jvg{Make it clear what do you mean by that? You mean that it may be the low framerate and a higher framerate could solve? Always make conclusions explicit.}\done{?}

On the MAHNOB-HCI dataset, we see that similar to the majority of signal processing methods, our method does not achieve a very good accuracy. 
An interesting observation is that the accuracy produced by almost all methods is close to that of a dummy baseline method that blindly predicts the mean heart rate of the dataset (\texttildelow71~bpm) for any input.
Apart from \cite{li2014remote}, only the deep learning based methods perform better.
This could be an indication that the high compression noise distorts the pulse information in the spatial averages of skin pixels.
Deep learning based methods are able to somewhat overcomes this, perhaps by learning to detect and filter out the spatial `pattern' of such compression noise.
In addition, deep learning methods might also be implicitly learning to track ballistic head movements of the subjects since it is also caused by blood pulsations~\cite{balakrishnan2013detecting,starr1939studies}.
In fact, the lowest error rate is obtained by the ballistic head movement based method described in \cite{haque2016}. 
This method does not rely on colour information at all, and hence its prediction is not affected by the underlying compression noise.
This suggests a similar conclusion: simple spatial averaging of pixel values is not sufficient to estimate HR accurately in this dataset due to the high compression noise.
%Similar error rates are obtained from the method described in \cite{haque2016}. This method estimates heart rates based on ballistic head movements and does not rely on colour information. 

\subsection{Heart Rate Variability Analysis}

\begin{figure}
	\floatbox[{\capbeside\thisfloatsetup{capbesideposition={left,bottom},capbesidewidth=0.48\textwidth}}]{figure}[\FBwidth]
	{\caption{Predicted vs ground truth heart rate variability (HRV) in terms of RMSSD (ms). 
	Each point represents one video in the dataset: (a) StableSet (\texttildelow2.5min duration); (b) VicarPPG  (\texttildelow1.5min duration); (c) PURE (\texttildelow1min duration).
	The estimated HRV shows fairly decent correlation with the ground truth.} %\done{write better caption}}
	\label{fig:HRVresults}\hspace{-0.6cm}}
	{\subfloat[][StableSet]{\includegraphics[width=.48\textwidth, trim={0 0 0 0.8cm}, clip]{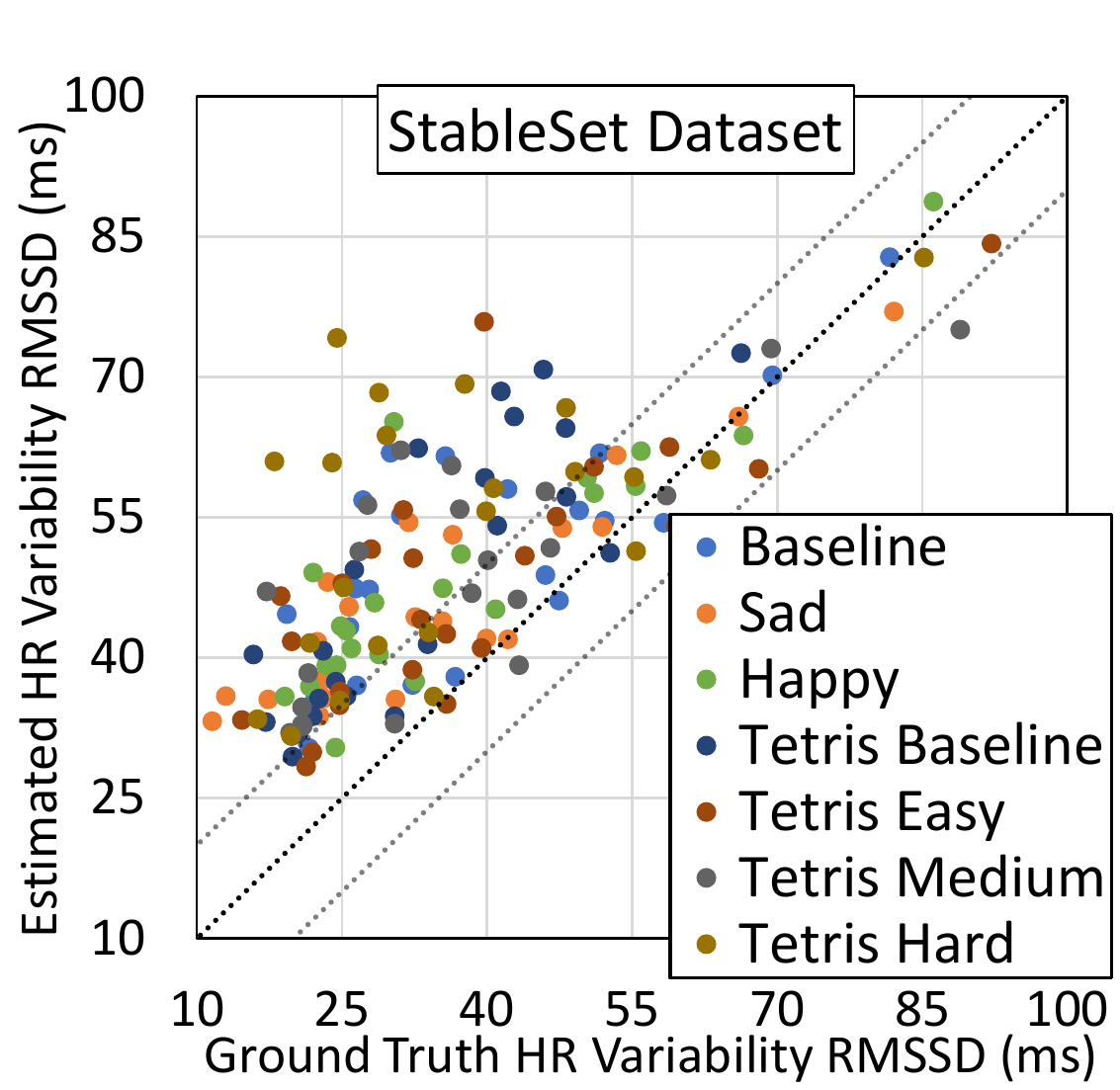}\label{HRVresults-StableSet}}}
	{\subfloat[][VicarPPG]{\includegraphics[width=.48\textwidth, trim={0 0 0 0.79cm}, clip]{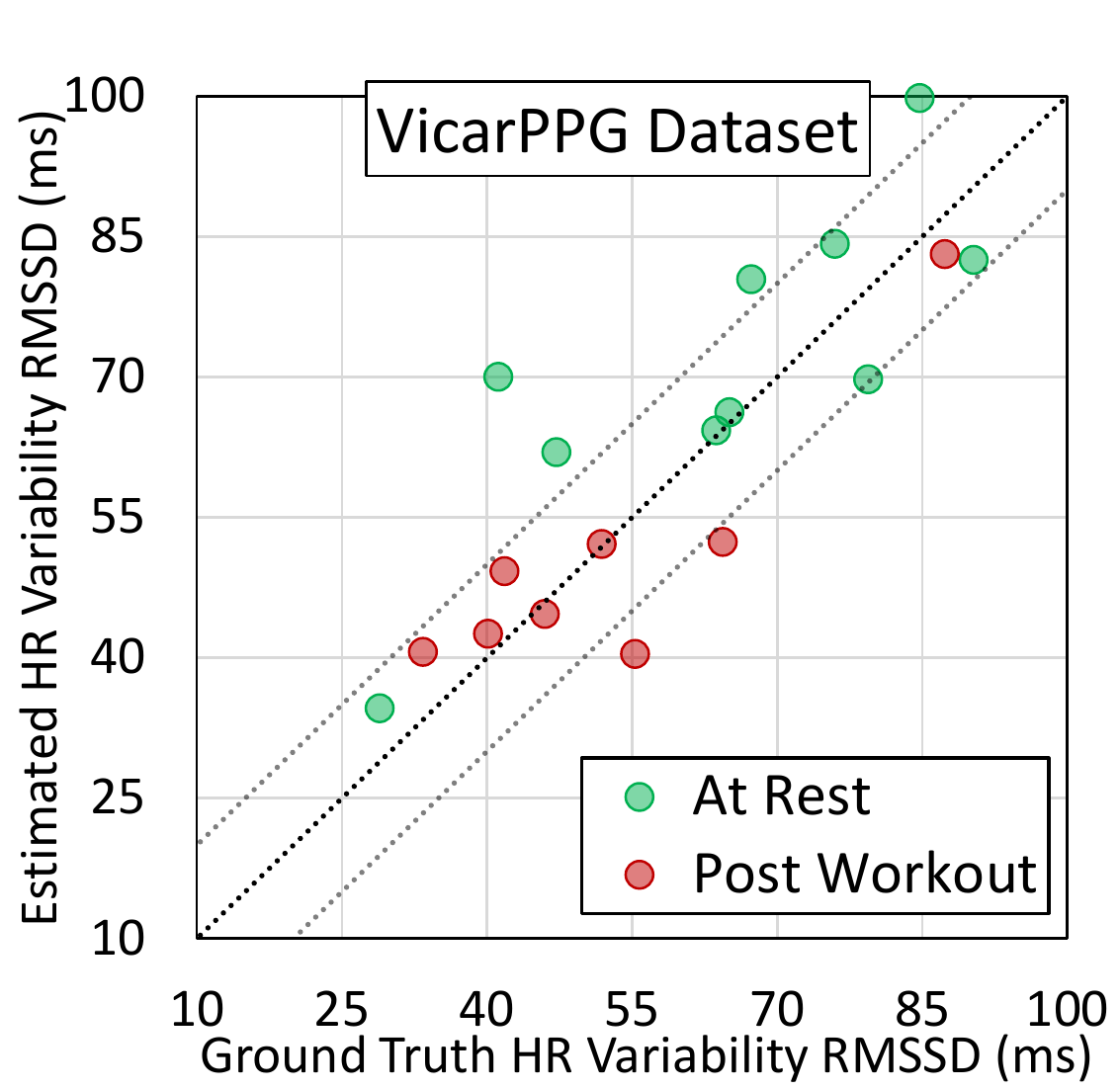}\label{HRVresults-VicarPPG}}}
	{\subfloat[][PURE]{\includegraphics[width=.48\textwidth, trim={0 0 0 0.79cm}, clip]{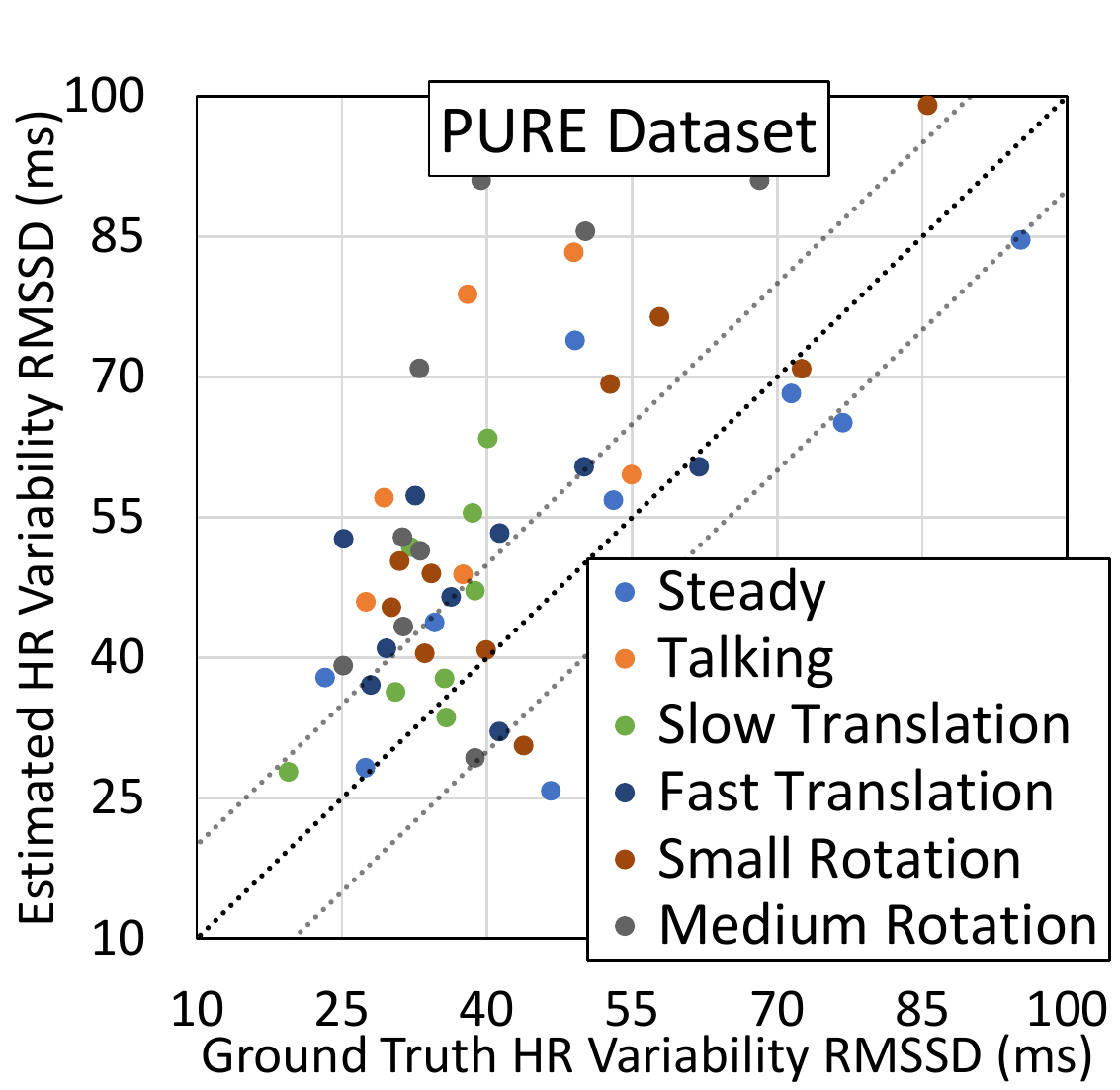}\label{HRVresults-PURE}}}
\end{figure}

% \begin{figure}
% 	\centering
% 	\subfloat[][]{\includegraphics[width=.48\textwidth, trim={0 0 0 0.8cm}, clip]{HRVresults-stableset.pdf}\label{HRVresults-StableSet}}\\
% 	\subfloat[][]{\includegraphics[width=.48\textwidth, trim={0 0 0 0.79cm}, clip]{HRVresults-vicarppg.pdf}\label{HRVresults-VicarPPG}}
% 	\subfloat[][]{\includegraphics[width=.48\textwidth, trim={0 0 0 0.79cm}, clip]{HRVresults-pure.pdf}\label{HRVresults-PURE}}
% 	%\subfloat[][]{\includegraphics[width=.4\textwidth]{HRVresults-MAHNOB.png}\label{HRVresults-MAHNOB}}
% 	\caption{Scatter plot of predicted vs ground truth heart rate variability in terms of RMSSD (ms) in each dataset. 
% 	Each point represents one video in the dataset: (a) StableSet (\texttildelow2.5min duration); (b) VicarPPG  (\texttildelow1.5min duration); (c) PURE (\texttildelow1min duration).
% 	Good correlation can be seen between the ground truth and estimated heart rate variability on all datasets.}
% 	%\jvg{too small} \jvg{why all in gray? Keep format consistent} 
% 	%\jvg{Conclusion missing. what do you want me to see? A reader first goes in comic book mode: add a conclusion to each caption}
% 	\label{fig:HRVresults}
%  \end{figure}

% \begin{figure}
% 	\centering
% 	\fbox{\includegraphics[width=\textwidth]{HRVresults-plot-dummy.jpg}}
% 	\caption{Scatter plots of HRV performance on all dataset except MAHNOB?}
% 	\label{fig:HRVresults}
% 	%\vspace{-8pt}
% \end{figure}

\begin{figure}
	\centering
	\includegraphics[width=0.8\textwidth, trim={0 0.65cm 0 0 }, clip]{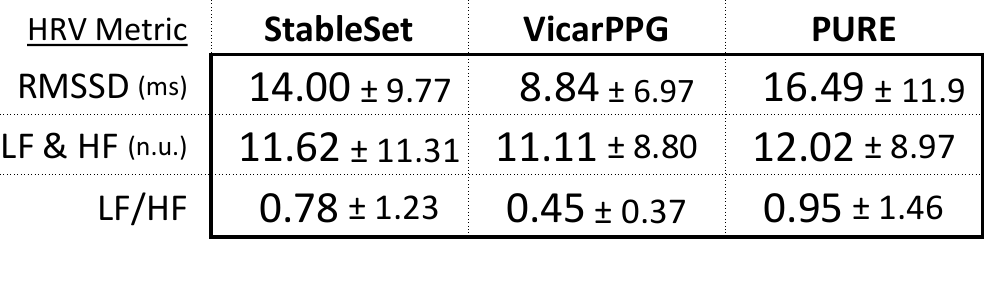}
	\captionof{table}{Heart rate variability computation performance of the proposed method in terms of mean absolute error. Based on the average human heart rate variability range, we see that good accuracies are obtained on all dataset.
	\ifsqueeze\vspace{-0.3cm}\fi}
	\label{tab:HRVresults}
	%\vspace{-8pt}
\end{figure}

The task of assessing heart rate variability is greatly more noise-sensitive than estimating heart rate. 
To validate our method on this task, we compute mean absolute error between the time-domain HRV measure \textit{root mean square of successive differences}~(RMSSD) of the predicted heart beats in comparison the ground truth. 
In addition, we also report the mean absolute errors of frequency domain metrics: Low Frequency (LF), High Frequency (HF) (in normalized units), and their ratio LF/HF.
%Section \ref{sec:hrandhrv} \hl{describes in more detail how these measures are calculated.}
%\todo{Define these terms! Or at least cite something here.}
%\jvg{In order is not in order}
%To validate our method on this task, we compute the time-domain HRV measure RMSSD using the predicted heat beats from our method and compare it to the RMSSD obtained from ground truth.
%We select RMSSD as our HRV metric because it is \hl{one of} the most widely used HRV measure in medical, psychological, and physiological analysis.
%In addition, we also report some frequency domain metrics like LF, HF and LF/HF to make cross-dataset comparisons to prior work easier.
%Unlike heart rate which is a short-term measure, HRV is typically computed over longer lengths of time. 
% in its ultra-short form is computed over one minutes at least. \mb{Not true ultra-short can be sub minute}
%\hl{Therefore, we evaluate HRV only on datasets that have videos of at least one minute in duration.} 
We evaluate HRV on the StableSet, VicarPPG and PURE datasets, all of which contain videos longer than one minute in duration. 
%\mb{Maybe we can say something along the lines, that with our HR result all reported HRV values would be nonsense}

The results of this analysis can be seen in Table \ref{tab:HRVresults} and Figure \ref{fig:HRVresults}. 
Similar to the results of HR analysis, our method predicts HRV with a good degree of accuracy on all three datasets over the length of the full video (1~min to 2.5~min).
Based on HRV literature~\cite{shaffer2017overview} and considering that the average human heart rate variability is in the range of [19-75]~ms RMSSD, error rates less than \texttildelow20~ms RMSSD can be considered acceptably accurate.
%\mb{Can we say that?}

\subsection{In-depth Analysis}
%\jvg{good}

\begin{figure*}
	\centering
	\includegraphics[width=0.9\textwidth, trim={0 0.23cm 0.1cm 0}, clip]{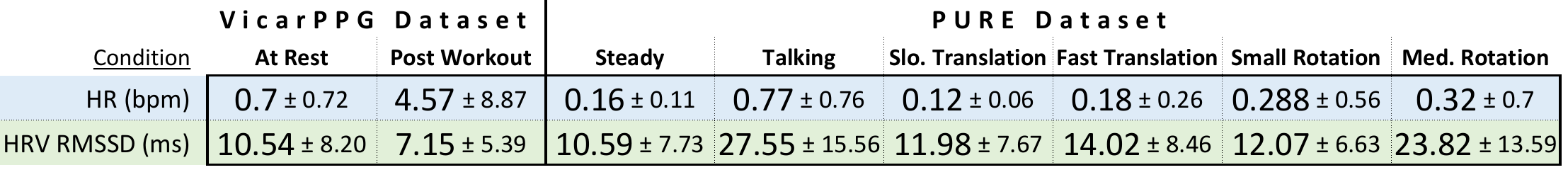}
	\captionof{table}{In-depth results of heart rate and heart rate variability analysis on VicarPPG and PURE datasets for all condition. Good accuracy is obtained in all movement conditions in the PURE dataset. \textit{Talking} performs relatively worse, likely due to incorrect face modelling caused by non-trivial facial deformations. The relatively lower performance in the post workout condition of VicarPPG is due to low frame rate artefacts in the video.
	%\ifsqueeze\vspace{-0.1cm}\fi
	}
	\label{tab:indepthresults}
	%\vspace{-8pt}
\end{figure*}

\begin{figure}
	\centering
	%\subfloat[][]{\includegraphics[width=.178\textwidth]{HRV-indepth-vicarppg.pdf}\label{HRVresults-indepth-StableSet}}
	\includegraphics[width=\textwidth]{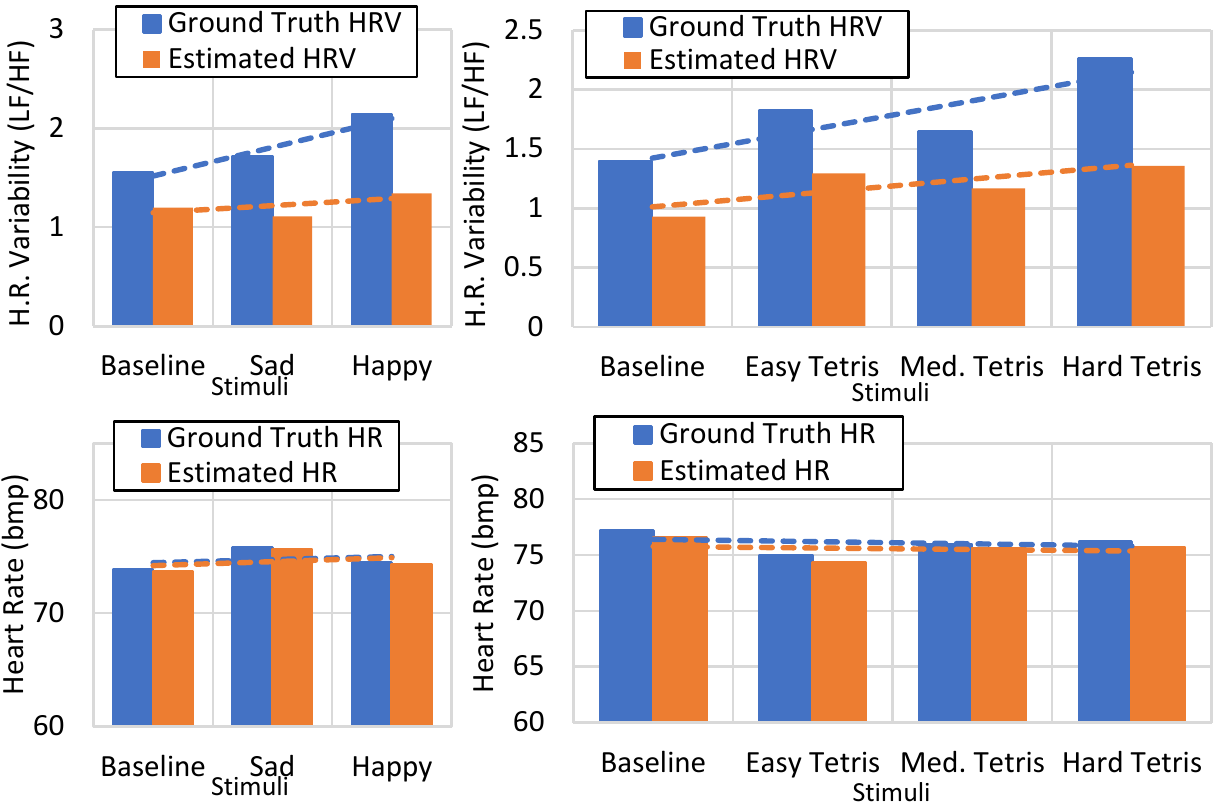}
	\caption{Comparison of HRV measure (LF/HF) for all conditions of the StableSet dataset. While the heart rates across all conditions remain constant, HRV is higher for emotional video stimuli and difficult Tetris tasks relative to baselines. This shows the utility of HRV over HR for measuring psychological conditions like stress.}
	%\ifsqueeze\vspace{-0.2cm}\fi}
	\label{fig:HRVresults-indepth}
 \end{figure}

Since both the VicarPPG and PURE dataset have physiological and physical condition partitions, it is possible to perform a more in-depth analysis of the HR and HRV results from our experiments. These in-depth individual error rates per condition can be seen in Table \ref{tab:indepthresults}. 

\paragraph{VicarPPG Conditions} 
It can be noticed that while the proposed method is fairly accurate over all conditions, it estimates heart rates in the post-workout condition less accurately than in the rest condition. 
This can also be observed in the scatter plot of Figure~\ref{fig:HRresults-VicarPPG}: while the overall average HR is accurate, there are more outliers in the higher HR region.
On closer examination, we found the primary reason for this to be that a much larger number of video segments with a very low variable frame rate were present in the post-workout partition.
Such low frame rate artefacts affect the estimation of higher HRs more severely as the Nyquist frequency requirement is also higher.
Figure~\ref{fig:HRexamples} also shows this for an example video from the VicarPPG dataset: the predicted HR follows the ground truth very well except for in the starting segment which has a higher heart rate (but low frame rate). % where the predictions are severely underestimated. 
If these videos are excluded (\texttt{06\_WORKOUT}, \texttt{08\_WORKOUT}), the error rates drop significantly to \textbf{0.99{\raisebox{.2ex}{$\scriptstyle\pm$}}0.8~bpm}, and so does the discrepancy between the two conditions.
%This region has a low variable frame rate. \mb{We could add something here about: lower framerate -> lower Nyquist -> not possible to measure elevated heart rates i.e. post workout}

\paragraph{PURE Conditions}
The PURE dataset conditions are based on the amount and type of movement the subjects perform.
As can be seen in Table~\ref{tab:indepthresults}, our method performs almost equally well in all movement conditions.
In fact, even large rotations and fast translations of the face are handled just as good as the \textit{Steady} condition.
This is primarily due to the motion noise suppression step and the translation-robustness of the appearance modelling. We are still able to closely track our regions of interest through the movement.
The worse performing condition is Talking.
This is somewhat expected as moving the mouth and jaw deforms the face in a non-trivial manner which the appearance model is unable to adapt well to. In addition, repetitive muscle movements in the face during talking can interfere with the observed colour changes.

%\vspace{-2.4cm}
\paragraph{StableSet Conditions}
%\ag{Work in progress. Not sure if we should include this section.}
%\todo{Include this only if any interested HRV results are obtained from StableSet}
Unlike the previous datasets, the conditions in the StableSet do not relate to physical (or physically induced) differences, but to the type of stimuli applied to the subjects while being recorded. These include two emotion-inducing videos (sad and happy), and three stress-inducing Tetris-game playing tasks with increasing levels of difficulty.

The average HR and HRV estimations in comparison with the ground truth for each condition is shown in Figure \ref{fig:HRVresults-indepth}. 
%As can be seen, 
While no significant differences can be seen in the average heart rate measurements of the subjects under different conditions, their HRV measurements show some interesting results:
the emotional videos and the higher difficulty Tetris games induce a higher HRV in the subjects when compared to the baseline.
These results demonstrate the usefulness of heart rate variability in assessing the underlying psychological conditions: HRV is able to highlight differences by showing variations under different conditions while HR stays the same.

\subsection{Processing Speeds}
\begin{figure}%[H]
	\centering
	\includegraphics[width=\textwidth, trim={0 0.88cm 0.1cm 0}, clip]{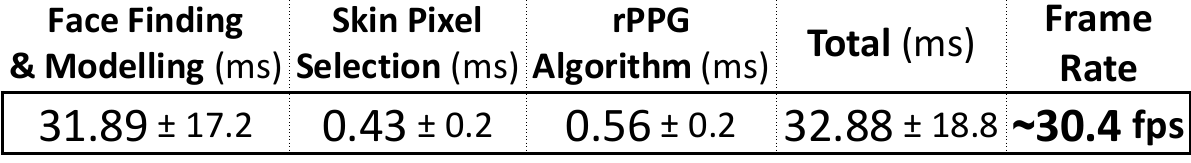}
	\captionof{table}{The processing speed of individual components of the proposed pipeline and the total frame rate for an input video (640$\times$480 resolution). Note that the bottleneck in the pipeline is face finding and modelling, while the rest require negligible time.
	%\ifsqueeze\vspace{-0.1cm}\fi
	}%\hl{, which can be swapped with a faster facial landmark detector}.}
	\label{tab:computation}
	%\vspace{-1cm}
\end{figure}
%\ag{Work in progress. Not sure if we should include this section.}
% For real time application, processing speed is just as vital as prediction accuracy. 
% The average CPU processing times of our method and its individual components is listed in Table \todo{make table?}.
% We see that on average this method is comfortably able to perform the full analysis with real-time speeds between 15 - 30~fps.
% At the same time, the accuracy of our system is high enough to compete with recent deep learning based methods, which are typically much heavier... \todo{how heavy?}
%Investigate: what nice properties does out method have: speed. 
%Average times: FaceAnalysis 54.77 pm 10.33 ms (Assuming a faster computer can run it at approximately 40% faster than mine)
%				Algorithm components = 2.187
%				Skin-pixel selection  = 0.79
%Intel-CORE i7 860

For real time application, processing speed is just as vital as prediction accuracy.
The average CPU processing times of our method and its individual components are listed in Table \ref{tab:computation} (on an Intel Xeon E5-1620).
We see that the method is comfortably able to perform the full analysis with a good real-time speed for a video resolution of 640$\times$480. 
For videos at 1280$\times$720, the analysis speed drops to \texttildelow23~fps, which can still be considered real-time. The only bottleneck in the pipeline is the face finding and modelling step, which is modular w.r.t the rPPG pipeline. Swapping this for faster face and landmark detectors (e.g. \cite{zhang2014facial} \texttildelow10~ms on CPU) can significantly improve overall processing speeds.
%

%We see that on average this method is comfortably able to perform the full analysis with real-time speeds at 17fps.
%The Face analysis step includes next to the described face detection and AAM fitting additional computational steps, which are irrelevant for the presented method.
%It can easily be identified as the bottleneck in the method allowing for a maximum of 18fps due to its execution time of 54ms.
%The remaining two steps of extracting the skin pixels from the ROI (0.79 ms) and processing the colour traces (2.18 ms) are fast enough to run the algorithm comfortably in real-time.
%The shown accuracy of our system is high enough to measure heart rate and heart rate variability at real-time speed.
%Dropping the additional computation steps from the face analysis or use of an alternative face finding algorithm could further improve the processing time of the presented method.

%At the same time, the accuracy of our system is high enough to compete with recent deep learning based methods, which are typically much heavier... \todo{how heavy?}

%% file: conclusion.tex
\section{Discussion}
\label{sec:disc}
We were able to obtain successful and promising results from our appearance modelling and signal-processing based rPPG method.
The results show that this method is able to obtain high accuracies, surpassing the state-of-the-art on two public datasets (VicarPPG and PURE).
We showed that small/large movements and rapidly changing heart rate conditions do not degrade performance. 
This can be attributed to the appearance modelling and noise suppression steps in the pipeline.
Only non-trivial facial deformations (e.g. during talking) proved slightly challenging, but the method still produced sub-1~bpm error rates in these conditions.

This high accuracy was obtain while being efficient: the method's computational costs were very low and the full pipeline could be executed in real-time on CPUs.
This is in contrast to deep learning based methods where larger models that can potentially form computational bottlenecks in such pipelines.
This efficiency is attained by taking advantage of our prior domain knowledge about the rPPG process, which the deep learning methods have to spend computational resources to learn and execute.

This high precision in estimating the pulse signal enables the measurement of heart rate variability (HRV), whose computation is sensitive to noise. 
HRV is a useful measure: As shown in the results, it can indicate underlying physio/psychological conditions (like stress) where HR is unable to show any difference.

A limitation of this method was observed in analysis of videos with very high compression rates. 
The resultant noise distorts the pulse signal almost completely when employing spatial averaging techniques. 
Deep learning methods like HR-CNN~\cite{vspetlikvisual} have shown better results in this setting, while it fails to match our method in cases with lower compression.
This could be because the network is able to learn the spatial patterns of this compression noise and filter it out, as well as track ballistic head movements and infer heart rate from it. 
In contrast, in lower compression cases, our prior domain knowledge assumptions perform more accurately.
% and efficiently
%In addition, the network might also be learning to track ballistic head movements and using that to predict heart rate.
While this makes our method well suited for modern videos, deep learning might be better suited for processing archival videos, often subjected to higher compression.

\section{Conclusion}
\label{sec:conc}
%\ifsqueeze\vspace{-0.1cm}\fi
This paper demonstrates a refined and efficient appearance modelling and signal processing based pipeline for remote photo-plethysmography. 
This method is able to estimate both heart rate and heart rate variability using cameras at real-time speeds.
This method was validated on multiple public datasets and state-of-the-art results were obtained on VicarPPG and PURE datasets.
%We validate this method on multiple public datasets and obtain state-of-the-art results on VicarPPG and PURE datasets.
We verify that this method is able to perform equally well under varying movement and physiological conditions.
However, while the estimations are precise under ordinary video-compression conditions, high levels of compression noise degrades the accuracy.